\documentclass[10pt,twocolumn]{article}
\usepackage[T1]{fontenc}

\usepackage{cite}

\usepackage{amsmath}

\usepackage[cmintegrals]{newtxmath}

\usepackage{url}

\usepackage{mdwmath}
\usepackage{mdwtab}

\usepackage{enumerate}
\usepackage{multirow} 
\usepackage[justification=centering]{caption}
\usepackage[colorlinks, citecolor=blue]{hyperref}
\usepackage{indentfirst}
\usepackage{array}
\usepackage{ragged2e}
\usepackage{enumitem}
\usepackage{comment}
\usepackage{tabularx}
\usepackage{graphicx}
\usepackage{multicol} 

\newcommand{\tabincell}[2]{\begin{tabular}{@{}#1@{}}#2\end{tabular}}

\usepackage{balance}

\usepackage{wrapfig}
\usepackage{geometry}
\usepackage{float}

\begin{document}
%
\title{A Systematic Literature Review on Federated Learning: From A Model Quality Perspective}

\author{Yi Liu	\thanks{Y. Liu is with the School of Computer Science and Engineering, Beihang University, Beijing, China. E-mail: zy1906505@buaa.edu.cn. },
	Li Zhang\thanks{L. Zhang is with the School of Computer Science and Engineering, Beihang University, Beijing, China. E-mail: lily@buaa.edu.cn. },
	Ning Ge\thanks{N. Ge is the corresponding author. She is with the School of Software, Beihang University, Beijing, China. E-mail: gening@buaa.edu.cn},
	Guanghao Li\thanks{G. Li was with the School of Software, Beihang University, Beijing, China. He is now with University of California, San Diego. E-mail: liguanghao@buaa.edu.cn. }}

%
\maketitle
\begin{abstract}
	\justifying
As an emerging technique, Federated Learning (FL) can jointly train a global model with the data remaining locally, which effectively solves the problem of data privacy protection through the encryption mechanism. The clients train their local model, and the server aggregates models until convergence. In this process, the server uses an incentive mechanism to encourage clients to contribute high-quality and large-volume data to improve the global model. Although some works have applied FL to the Internet of Things (IoT), medicine, manufacturing, etc., the application of FL is still in its infancy, and many related issues need to be solved. Improving the quality of FL models is one of the current research hotspots and challenging tasks. This paper systematically reviews and objectively analyzes the approaches to improving the quality of FL models. We are also interested in the research and application trends of FL and the effect comparison between FL and non-FL because the practitioners usually worry that achieving privacy protection needs compromising learning quality. We use a systematic review method to analyze 147 latest articles related to FL. This review provides useful information and insights to both academia and practitioners from the industry. We investigate research questions about academic research and industrial application trends of FL, essential factors affecting the quality of FL models, and compare FL and non-FL algorithms in terms of learning quality. Based on our review's conclusion, we give some suggestions for improving the FL model quality. Finally, we propose an FL application framework for practitioners.
\end{abstract}
	


%

\section{Introduction}
%
%
%
%
Application need for machine learning is highly increasing in many fields like Internet of Things (IoT), natural language understanding, machine vision, pattern recognition, etc. The training process of machine learning requires a large number of shared data samples. Since 2018, data privacy protection has started to restrict the application of machine learning. The European Union issued the General Data Protection Regulation (GDPR)\cite{EU2018}. McMahan et al.\cite{DBLP:conf/aistats/McMahanMRHA17} proposed a federated learning (FL) method in 2016 to solve the conflict between data privacy protection and machine learning training. This method coordinates the device’s operation through the central server under the premise that the data is stored locally. As defined by Kairouz P et al.\cite{kairouz2019advances}, \emph{“Federated learning is a machine learning setting where multiple entities (clients) collaborate in solving a machine learning problem, under the coordination of a central server or service provider.  Each client’s raw data is stored locally and not exchanged or transferred; instead, focused updates intended for immediate aggregation are used to achieve the learning objective.”}

\par
Compared to traditional machine learning, the FL algorithm consists of three major parts, i.e., the learning algorithm and training method, the data privacy protection mechanism, and the participant incentive mechanism. Learning algorithm and training method refers to the server’s iterative process of aggregation after each client trains the model locally. It uses a privacy protection mechanism to protect data privacy and uses an incentive mechanism to encourage the clients to participate in the joint model training. The quality of an FL algorithm is thus closely related to three aspects: (1) the quality of the learning model and training, (2) the quality of the federated privacy protection mechanism, and (3) the quality of the incentive mechanism.

\textbf{Quality of learning model design and training.} An FL algorithm is designed in two parts. One is for the client-side local learning, and the other is for the server-side aggregation. The learning quality measured using the accuracy, the precision, etc., highly depends on the model design and training between the client and server sides. 

\par
\textbf{Quality of federated privacy protection.} The training process of FL can keep the data locally, which improves the quality of privacy protection compared to centralized machine learning. To protect data privacy, the training process of FL also needs to ensure the following two facts. (1) The trained model transmitted during the training process contains information on the original data. It thus requires avoiding inferring the original data from the trained model. (2) The server-side only obtains the agreed information from the client-side and does not additionally get other redundant information. It thus requires making sure that only intermediate results are sent without additional information. 

\par

\textbf{Quality driven by insensitive mechanism.} The effectiveness of machine learning also depends on the quality and quantity of data used for training. Although FL ensures that the data remain locally, the clients need to provide their resources, such as computing power, data samples, communication cost, etc. These factors may cause clients to be unwilling to participate in FL without compensation. In FL, the clients with high-quality and large-volume data often cannot obtain higher returns than training alone, and the clients with a small amount of data are more interested in participation. Thus, it is necessary to use some incentive mechanisms to maximize the overall benefits, ensure that individual benefits are not harmed, and promote more clients with high-quality data to participate.

\par

Since 2019, FL has been paid more and more attention from the public. Google is the first to apply FL to practical applications, namely, keyboard next word prediction\cite{hard2018federated}. Many works are starting to apply FL to solve the joint learning problem under the data island. As a result, many hot research issues of FL have emerged. Some high-quality review articles are published  \cite{jin2020survey,kairouz2019advances,DBLP:journals/spm/LiSTS20,DBLP:journals/cem/LiSM20,DBLP:journals/corr/abs-1911-06270,DBLP:WeiFederated2020IEEE,DBLP:journals/corr/abs-2003-08119,DBLP:journals/corr/abs-2003-02133,niknam2019federated,DBLP:journals/network/MaLDYSQP20,DBLP:journals/network/DongCHL19}. These articles summarized application problems and research trends of FL. Some focus on the privacy protection problem, the emerging algorithms of FL, the FL data/device heterogeneity, communication efficiency/stability, or the traceability issues. Most practitioners expect a general application framework to ensure high-quality FL models. However, there is no literature review to give a systematic answer to this question. Our review focuses on the FL model quality. We investigate and analyze the existing literature from the learning model design and training, the privacy protection mechanism, and the incentive mechanism. We are also interested in the difference between centralized machine learning and FL under the same data set. With this aim, we summarize five research questions. Note that to focus on the major concerns in FL, the data/equipment heterogeneity, communication efficiency/stability, and traceability issues of FL are out of our review’s scope. Nevertheless, they can also affect the quality of the FL model. 

We are also interested in the research and application trends of FL and the effect comparison between FL and non-FL because the practitioners usually worry that achieving privacy protection needs compromising learning quality.

\par

Based on the above analysis, this review conducts investigations from the following five research questions (RQs):
\noindent
\begin{itemize}[leftmargin=*]
	\item \textbf{RQ1}: Since FL was proposed in 2016, what is the overall research and application trends of FL? 
	\item \textbf{RQ2}: What methods to improve the quality of FL model design and training? 
	\item \textbf{RQ3}: What methods to improve the quality of data privacy protection in FL?
	\item \textbf{RQ4}: What are the incentive mechanisms for improving the data quality of FL?
	\item \textbf{RQ5}: Can FL achieve similar learning effects to non-FL on the same dataset? What are the conditions when FL and non-FL results are similar? 
\end{itemize}

In this work, we first analyze the main factors that influence FL models’ quality. Then, we propose an FL application framework to help ensure model quality. The main contributions of this article are twofold:

\begin{enumerate}[leftmargin=*]
	\item We conducted a systematic and objective analysis of 147 articles. By conducting statistical analysis on the overall situation, algorithms, encryption, incentive mechanism, and effects of FL, we finally answered the five predefined RQs.
	\item After summarized the influencing factors of the FL model quality, we proposed an FL application framework with detailed technique options. We hope this framework can be helpful for the practitioners of FL.
\end{enumerate}

This paper’s structure is as follows. Section 2 introduces some key concepts in FL and designs the implementation steps of the review. Section 3 analyzes the results of our review. Section 4 proposes an application framework for ensuring the FL model quality. Section 5 discusses the threat to validity. Section 6 concludes this work.

\section{Systematic Literature Review}
\subsection{Preliminaries of Federated Learning}
\begin{figure}[htbp]
	\centering
	\includegraphics[width=3.4in]{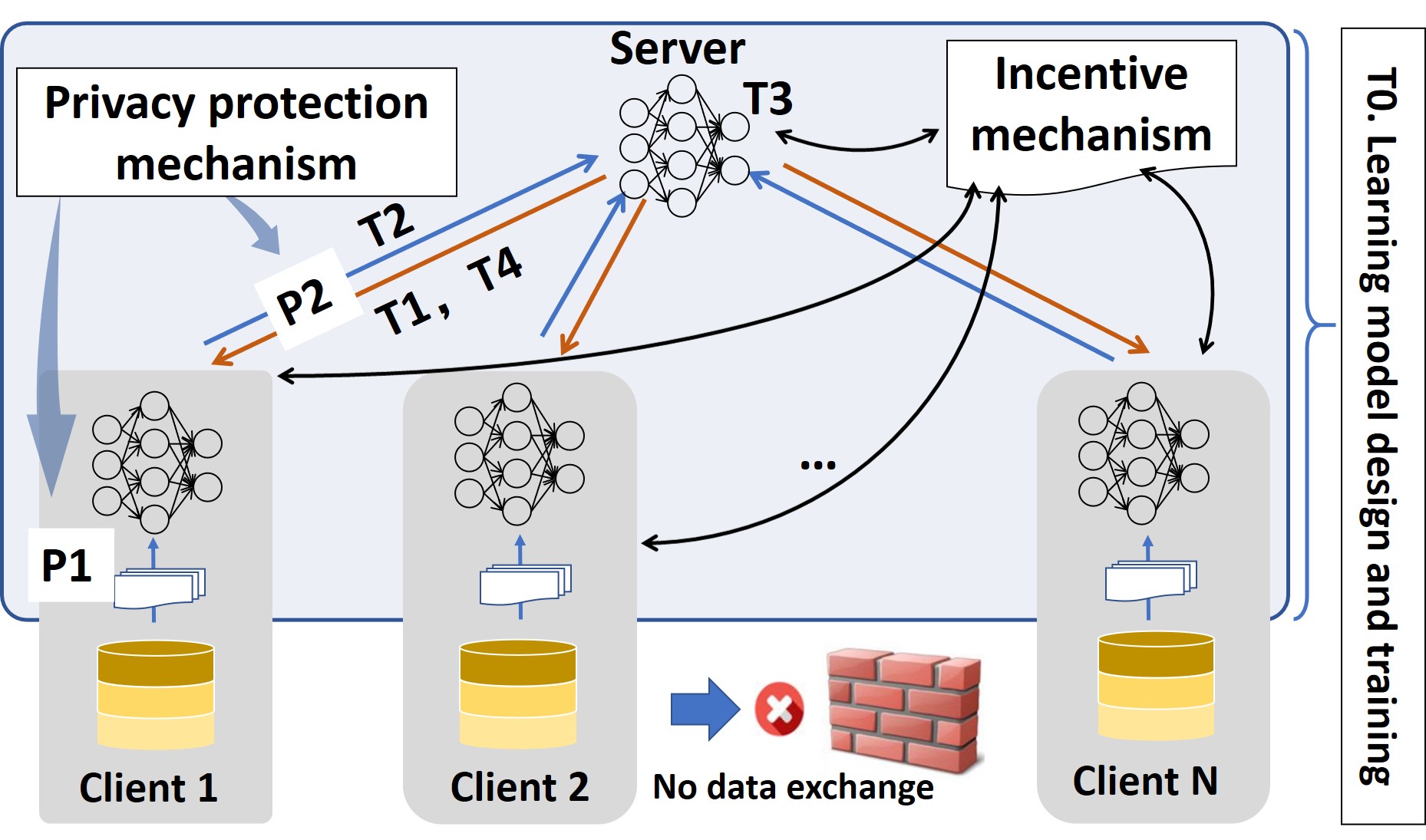}
	\caption{Federated learning process \cite{DBLP:journals/tist/YangLCT19}}
	\label{fig_2-1}
\end{figure}

FL is used to establish a learning model based on distributed data sets \cite{Y2019federatedLearning}. As shown in Fig. \ref{fig_2-1}, the training process of FL consists of three major parts: learning algorithm and training method, privacy protection mechanism, and incentive mechanism. We explain each part hereafter.

\par

\par
\textbf{(1) Model design and training}
\par
In the training process, the server trains the learning model by repeating the following steps until reaching the termination condition:
\begin{itemize}[leftmargin=*]
	\item \textbf{T0: Design the client-side and server-side learning algorithm.} The practitioner designs the learning algorithm for the clients and server regarding the application requirements.
	\item \textbf{T1: The server selects clients and delivers the model.} The server samples from a group of clients that meet the qualification requirements. Then, the server sends to the clients chosen a shared model and training setting information.
	\item \textbf{T2: Client model update.} The selected client updates the model by executing the training program.
	\item \textbf{T3: Server aggregation.} The server aggregates according to the client’s model or the parameters sent from the client’s model.
	\item \textbf{T4: Server model updates.} The server updates the shared model based on the aggregation results in the current round.
\end{itemize}

\textbf{(2) Privacy protection mechanism}
\par
The article \cite{Quantification2019federated} discusses the privacy leakage problem based on a federated approximated logistic regression model. It shows that such a gradient’s leakage could leak the complete training data if all inputs are 0 or 1. The article \cite{InvertingGradients2020federated} proved that sharing parameter gradients is not safe. According to the image’s parameter gradients, the author used a cosine similarity loss and optimization methods from adversarial attacks to accurately reconstruct high-resolution images. The author shows numerically that even averaging gradients over several iterations or several images does not protect the user’s privacy in FL applications in computer vision. FL can protect user data privacy from the following two aspects:
\begin{itemize}[leftmargin=*]
	\item \textbf{P1: Data encryption training.} It avoids inferring the original data from the model.
	\item \textbf{P2: Encryption of the data transmission process.} It achieves secure multi-party calculations to ensure that only the intermediate results without additional information are transmitted during the model training process.
\end{itemize}

\textbf{(3) Incentive mechanism}
\par
It is necessary to provide sufficient incentives for participants. The incentive mechanism aims to fairly and justly share the FL’s profits, stimulating the participants to continuously participate in the FL. Simultaneously, the incentive mechanism can also prevent malicious participants from controlling the federated learning process.
\par
\begin{figure}[htbp]
	\centering
	\includegraphics[width=3.4in]{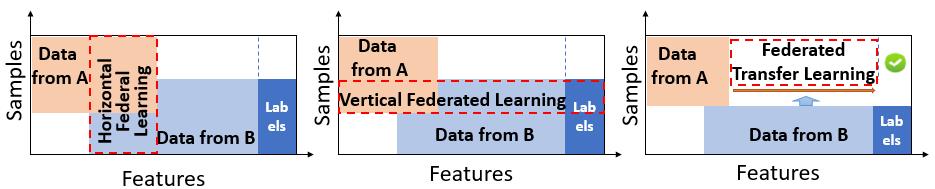}
	\caption{Three federated settings \cite{DBLP:journals/tist/YangLCT19}}
	\label{fig_2-2}
\end{figure}

There are three major types of federated settings regarding the data features in FL as shown in Fig. \ref{fig_2-2} \cite{Y2019federatedLearning}, i.e., horizontal federated learning, vertical federated learning, and federated transfer learning. 
\begin{itemize}[leftmargin=*]
	\item \textbf{Horizontal Federated Learning}: two datasets share the same feature space but differ in sample ID space. For example, two banks in different regions have tiny user intersection. But the business is similar, and the feature space is the same.
	\item \textbf{Vertical Federated Learning}: two datasets share the same sample ID space but differ in feature space. For example, banks and hospitals in the same city. Their user set may contain most of the area’s residents, so users’ intersection is frequent. But the business of banks and hospitals is different, so the feature space is different.
	\item \textbf{Federated Transfer Learning}: two datasets differ not only in samples but also in feature space. For example, a bank in China and a hospital in the United States. In this case, the sample space intersection and feature space intersection are both small.
\end{itemize}

\subsection{Systematic literature review}

This section shows a detailed introduction to the implementation process of a systematic literature review. Before conducting a systematic literature review, we develop a review plan. According to the articles \cite{libguides2020,wohlin2012experimentation,Kitchenham2007Guidelines}, a systematic review usually follows the same steps as shown hereafter:
\begin{enumerate}[leftmargin=*]
	\item Formulate research questions
	\item Develop and execute a search strategy
	\item Formulate selection criteria for literature
	\item Develop and execute literature selection procedures
	\item Formulate and execute literature extraction strategies
	\item Extract data and comprehensive analysis
\end{enumerate}

The following sections describe the detailed implementation of these steps. 

\subsubsection{Formulate research questions}

\begin{itemize}[leftmargin=*]
	\item \textbf{RQ1}: Since FL was proposed in 2016, what is its overall research and application trends? We divide this RQ into the following four sub-RQs:
	\begin{itemize}
		\item \textit{RQ1.1}: How hot is the research on FL?
		\item \textit{RQ1.2}: How about the publication trends of FL?
		\item \textit{RQ1.3}: What is the participation degree of academia and industry in FL research? What is the degree of participation of each country in FL?
		\item \textit{RQ1.4}: What are the application scenes of FL, and what practical problems are solved?
	\end{itemize}
	\item \textbf{RQ2}: What methods to improve the quality of FL learning algorithm and training method? We investigate the following five sub-RQs:
	\begin{itemize}
		\item \textit{RQ2.1}: Are the algorithms between clients the same? Are the client-side and server-side algorithms the same? Are there constraints on the client-side and server-side algorithms?
		\item \textit{RQ2.2}: Considering the client-side, what are the methods to improve the quality of FL?
		\item \textit{RQ2.}3: Considering the server-side, what are the aggregation algorithms to improve the quality of FL?
		\item \textit{RQ2.4}: Considering the server-side, what methods improve the aggregation algorithm to enhance training quality?
		\item \textit{RQ2.5}: Considering the server-side, what methods improve the training quality by enhancing the federated training process?
	\end{itemize}
	\item \textbf{RQ3}: What methods to improve the data privacy protection of FL?
	\item \textbf{RQ4}: What are the incentive mechanisms to improve the data quality of FL?
	\item \textbf{RQ5}: Can FL achieve similar learning effects to non-FL on the same dataset? What are the conditions when FL and non-FL results are similar?  
\end{itemize}

\subsubsection{Develop and execute a search strategy}

\textbf{(1) Databases for searching}
\par
The articles \cite{wohlin2012experimentation,Kitchenham2007Guidelines} mention some high-quality databases for searching basic research resources. Due to the rapid development of FL, many articles are available on platforms such as arXiv, which are semi-published. The Google Scholar database contains articles on the arXiv platform. We have collected both officially published documents and semi-published documents. We only obtain documents that have access methods and finally chose the following five databases:

\begin{itemize}[leftmargin=*]
	\item IEEE;
	\item ACM;
	\item Google Scholar;
	\item Science direct;
	\item Scopus.
\end{itemize}
\par

\textbf{(2) Retrieval time range}
\par
Considering that FL was proposed in 2016, and the systematic literature review began in April 2020, the final retrieval time range is from 2016 to March 2020.

\par
\textbf{(3) Search string}
\par
In the selected database, we first tried various search string criteria. The search string finally determined to obtain the most useful information is as follows:
\par
(FL) or (federated (deep or machine or reinforcement or transfer or active) learning) or (federated incentive) or (federated (application or framework)) or (federated (algorithm or method or approach)) or (federated model aggregation)

\par

\textbf{(4) Search results}
\par
The final search results are shown in Table \ref{table_2-1}.

\begin{table*}[htbp]
	\caption{Search result}
	\label{table_2-1}
	\centering
	\begin{tabular}{c|c|c|c|c|c}
		\hline
		\textbf{Google Scholar} & \textbf{ACM} & \textbf{IEEE} & \textbf{Science direct} & \textbf{Scopus} & \textbf{Total} \\ \hline
		680 & 11 & 186 & 26 & 237 & 1140 \\ \hline
	\end{tabular}
\end{table*}

\subsubsection{Formulate selection criteria for literature}

Based on the research content, the following exclusion criteria are formulated:
\begin{itemize}[leftmargin=*]
	\item Not related to FL;
	\item Non-English literature;
	\item Too short or incomplete;
	\item Not related to the research questions mentioned in Section 2.2.1.
\end{itemize}

\subsubsection{Develop and execute literature selection procedure}

We established a three-stage literature selection process. Each stage has different selection criteria, and each stage records the reasons for literature exclusion:
\begin{itemize}[leftmargin=*]
	\item Stage 1: From the 1140 papers in Section 2.2.2, we preliminary excluded non-FL papers and papers that are not related to the research questions by title and abstract. We excluded 698 papers in total, leaving 442 articles.
	\item Stage 2: From the 442 documents in stage 1, we excluded non-English paper, too short, or incomplete ones. We excluded 30 documents in total, leaving 412 articles.
	\item Stage 3: From the 412 documents in stage 2, we excluded papers that are not related to the research questions in Section 2.1, according to the literature’s content, and removed duplicate documents. We excluded 265 documents in total, leaving 147 articles.
\end{itemize}

\par

Table \ref{table_2-2} shows the results of each selection stage. In 11 of the 147 articles are reviews \cite{jin2020survey,kairouz2019advances,DBLP:journals/spm/LiSTS20,DBLP:journals/cem/LiSM20,DBLP:journals/corr/abs-1911-06270,DBLP:WeiFederated2020IEEE,DBLP:journals/corr/abs-2003-08119,DBLP:journals/corr/abs-2003-02133,niknam2019federated,DBLP:journals/network/MaLDYSQP20,DBLP:journals/network/DongCHL19}.  

\begin{table*}[htbp]
	\caption{Selection result}
	\label{table_2-2}
	\centering
	\begin{tabular}{c|c|c|c|c|c|c}
		\hline
		& \textbf{\tabincell{c}{Google Scholar}}& \textbf{ACM} & \textbf{IEEE} & \textbf{\tabincell{c}{Science direct}} & \textbf{Scopus} & \textbf{Total} \\ \hline
		\textbf{Stage 1} & 296                     & 6            & 25            & 7                       & 108             & 442            \\ \hline
		\textbf{Stage 2} & 274                     & 6            & 25            & 7                       & 100              & 412            \\ \hline
		\textbf{Stage 3} & 88                      & 6            & 13            & 4                       & 36              & 147            \\ \hline
	\end{tabular}
\end{table*}

\subsubsection{Formulate and execute literature extraction strategies}

After determining the literature selection procedure, we designed a data extraction form to record the literature’s information. According to the article\cite{zhang2018Investigation}, we obtain the data extraction form following the steps defined hereafter:
\begin{itemize}[leftmargin=*]
	\item Step 1: design a preliminary data extraction form based on the research questions.
	\item Step 2: randomly select 10 sample papers from the original ones.
	\item Step 3: fill in the data extraction form.
	\item Step 4: analyze the extracted data.
	\item Step 5: modify the data extraction form.
\end{itemize}

\par

Repeat steps 2 to 5 to form the final data extraction form, as shown in Table \ref{table_2-3}.

\begin{table*}[htbp]
	\caption{Data Extraction Form}
	\label{table_2-3}
	\centering
	\begin{tabular}{m{4.8cm}<{\centering}|m{10.5cm}<{\centering}|m{1.5cm}<{\centering}}
		\hline
		\textbf{Item}  & \textbf{Content}  & \textbf{Remark} \\ \hline
		Title  & Title of article  & \\ \hline
		Author  & Author's name  &  \\ \hline
		Type of article  & Journal/Conference/Workshop/Book/Phd thesis/arXiv/bioRxiv& \\ \hline
		Source and time of article  & Journal/Conference name, publication time   &  \\ \hline
		Author's unit & The university or institution to which the author belongs  & \\ \hline
		Country of research institution  & Country of research institution & \\ \hline
		Page range  & Page range   & \\ \hline
		URL of article  & URL of article &\\ \hline
		URL of Data set  & URL of Data set & \\ \hline
		URL of source code & URL of source code &\\ \hline
		Scenes& 1. Internet and Finance 2. Healthcare 3. Urban Computing and Smart City 4. Edge Computing and Internet of Things 5. Physical Information System 6. Industrial Manufacturing &  \\ \hline
		Problem solved  & The actual problem to be solved in the article  &  \\ \hline
		Federated setting  & 1. Horizontal(H) 2. Vertical(V) 3. Transfer(T) 4. Reinforcement  & Multiple \\ \hline
		Aggregation algorithm  & Design (improvement) goal, type, consideration, effect &  \\ \hline
		Machine learning algorithm  & Algorithm used for client training& \\ \hline
		Privacy  & Methods and effects of protecting privacy   & \\ \hline
		Incentive mechanism & Scenes, research questions, considerations, incentive goals, reward methods and effects &\\ \hline
		FL vs non-FL & 1. FL is better than non-FL 2. Similar (±1\%) 3. Non-FL is 1-5\% higher than FL 4. Non-FL is 5-10\% higher than FL   & Multiple        \\ \hline
		Improvements compared to traditional FL      & The purpose, method and effect of improvement     &   \\ \hline
	\end{tabular}
\end{table*}

\subsection{Reviews for Federated Learning}

We have collected 11 reviews \cite{jin2020survey,kairouz2019advances,DBLP:journals/spm/LiSTS20,DBLP:journals/cem/LiSM20,DBLP:journals/corr/abs-1911-06270,DBLP:WeiFederated2020IEEE,DBLP:journals/corr/abs-2003-08119,DBLP:journals/corr/abs-2003-02133,niknam2019federated,DBLP:journals/network/MaLDYSQP20,DBLP:journals/network/DongCHL19} related to FL.
\begin{itemize}[leftmargin=*]
	\item The literatures \cite{DBLP:journals/corr/abs-1911-06270,DBLP:WeiFederated2020IEEE,DBLP:journals/corr/abs-2003-08119} is biased towards specific areas, focusing on the statistical challenges, communication efficiency/stability issues, device/data heterogeneity issues, privacy and security issues, traceability and accountability in the process of combining mobile edge computing, healthcare and FL. They also give current solutions. The literature \cite{niknam2019federated} discusses several possible applications in 5G networks: edge computing and caching, spectrum management, 5G core network, and described the key technical challenges for future research of FL in the wireless communication environment: security and privacy challenges, algorithms Related challenges, challenges in the wireless environment.
	\item The literatures \cite{DBLP:journals/cem/LiSM20,DBLP:journals/corr/abs-2003-02133,DBLP:journals/network/MaLDYSQP20,DBLP:journals/network/DongCHL19} discuss data privacy protection in FL. Literature \cite{DBLP:journals/cem/LiSM20} discusses valuable attack mechanisms and proposes corresponding solutions to corresponding attacks. Literature \cite{DBLP:journals/corr/abs-2003-02133} introduces the basic knowledge and key technologies of various attacks and discusses the future research direction of achieving more robust privacy protection in FL. The literature \cite{DBLP:journals/network/MaLDYSQP20} discusses the protection of privacy and security when designing the FL system and divides the methods into three categories: client-side privacy protection, server-side privacy protection, and security protection for FL. The privacy and security issues of FL are divided into convergence, data poisoning, scaling up, and model aggregation problems. The author provides solutions to related problems. Literature \cite{DBLP:journals/network/DongCHL19} analyzes the federated security problems under Byzantine adversaries and divides the current security FL algorithms (SFLAs) into four categories: aggregation rule-based SFLAs, preprocessing based SFLAs, model-based SFLAs, and adversarial detection-based SFLAs. The author provided a qualitative comparison of current SFLAs and reviewed some typical work on SDLAs.
	\item The literature \cite{jin2020survey}briefly reviews semi-supervised algorithms and FL, and then analyzed federated semi-supervised learning, including settings and potential methods.
	\item The literature \cite{kairouz2019advances} comprehensively discusses the latest progress of FL and open issues, including the algorithms, efficiency, privacy issues, robustness, and fairness of FL.
	\item The literature \cite{DBLP:journals/spm/LiSTS20} discusses the four challenges in FL: expensive learning, system heterogeneity, statistical heterogeneity, and privacy issues. It also discusses related solutions to these four challenges.
\end{itemize}

\par

Existing review papers usually focus on the issues related to FL in a specific application scene, the privacy protection mechanism, and the emerging algorithms. Some works aim to solve the problems related to data/device heterogeneity, communication efficiency/stability, and traceability. How to follow a framework of FL that can guarantee the model quality is not yet studied. From the perspective of FL applications, this paper focuses on choosing an FL framework when using FL to solve a joint model training problem and suggests improving the quality of FL models. We focus on designing the federated learning process to train a model with high accuracy. From which factors that influence FL, we can improve FL.

\section{Results}

Through reading and analyzing articles related to FL in the five major databases, we conduct statistical analysis and present the results of a systematic literature review. These results correspond to the research questions raised in Section 2.2.1.

\subsection{RQ1: The overall situation of FL}

\subsubsection{RQ1.1 How hot is the research on FL? }
\begin{figure}[htbp]
	\centering
	\includegraphics[width=3.5in]{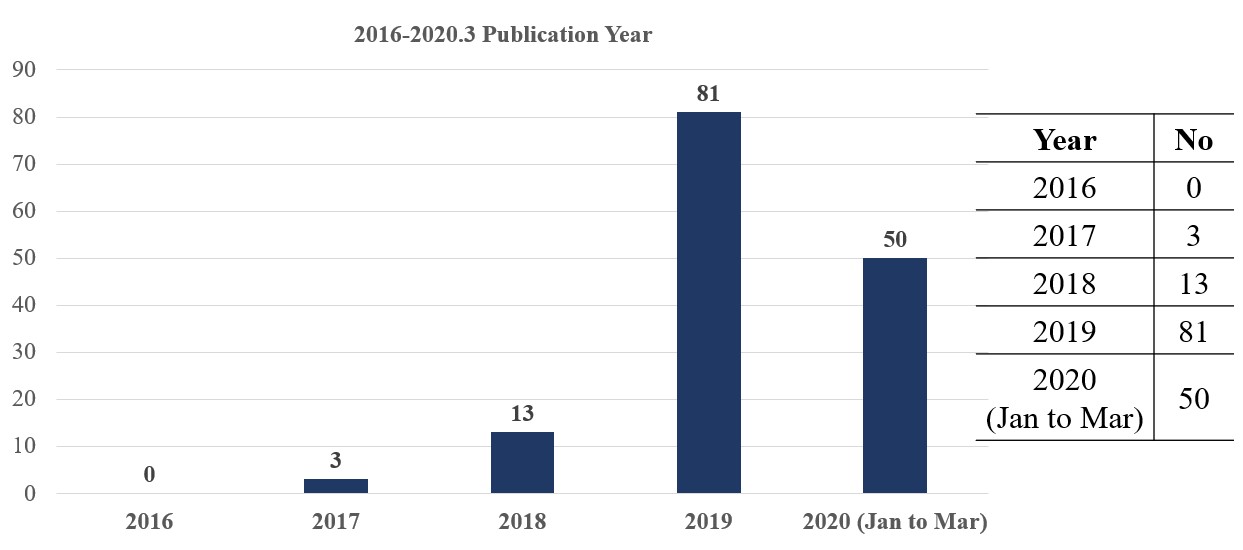}
	\caption{(2016-2020.3) Publication year of articles}
	\label{fig_3-1}
\end{figure}

To understand the heat of FL, we conducted statistics for the publication year of literature. Fig.\ref{fig_3-1} shows the statistical result. FL was first proposed in 2016. There were only 13 articles in 2018. However, by 2019, there were 81 articles published, and the number of articles published in the first three months of 2020 reached 50. Thus, the current development of FL is still in the rising stage.

\subsubsection{RQ1.2 How about the publication trends of FL? }
\begin{figure}[htbp]
	\centering
	\includegraphics[width=3in]{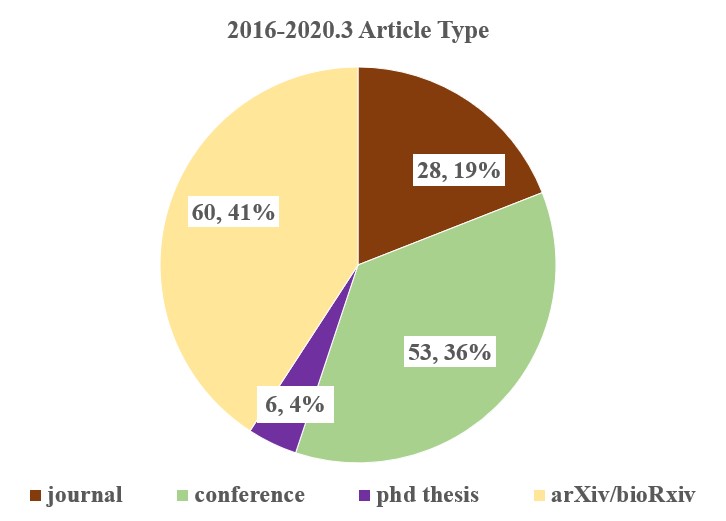}
	\caption{(2016-2020.3) Article type}
	\label{fig_3-2}
\end{figure}

In order to understand the publication trends of FL, we conducted statistics on the types of articles, as shown in Fig. \ref{fig_3-2}. During articles collection, there are fewer doctoral/master's theses in the selected database, resulting in a small proportion of the master's/doctoral theses. Most are semi-published works on arXiv/bioRxiv. Others are journal and conference papers. 

\par
As can be seen from Fig. \ref{fig_3-2}, as an emerging research direction, FL has many gaps in theory and application. Therefore, many research works have emerged in the short term and were published on arXiv /bioRxiv. Among them, some articles are under review in conferences or journals. Platforms such as arXiv are often used to develop fast new technologies. A total of 147 documents were counted, and 41\% of the articles were published on platforms such as arXiv, indicating the rapid development of FL. 36\% of the articles were published in the conferences, which also illustrates the new and fast characteristics of FL development. 19\% of the articles were published in journals, indicating that FL has produced more mature results. 4\% of the articles were master’s or doctoral theses, indicating that FL has been studied systematically.

\par
We further counted the published conferences of journals, and the articles related to FL covers a wide range, covering a wide range of 42 conferences and 21 journals. There are currently no conferences or journals dedicated to FL. There are no albums dedicated to FL. Many articles have been published in the IEEE International Conference on Big Data, IEEE Global Communications Conference, AAAI Conference on Artificial Intelligence, IEEE Internet of Things Journal, and IEEE Access.
\par
The publication form of FL literature shows that FL is a new hot topic. The wide variety of journals and conferences shows that FL is widely used in many research fields, such as artificial intelligence, communications, and big data. FL has broad application prospects.

\subsubsection{RQ1.3 What is the participation degree of academia and industry in FL research? What is the degree of participation of each country in FL?}

We conducted statistics for the sources of articles. We divided the sources into the following three categories:
\begin{itemize}[leftmargin=*]
	\item \textbf{Academic}, the authors only come from research institutions.
	\item \textbf{Industrial}Industrial, the authors only come from industrial units.
	\item \textbf{Combined}, the authors come from both research institutions and industrial units.
\end{itemize}

\par

\begin{figure}[htbp]
	\centering
	\includegraphics[width=3in]{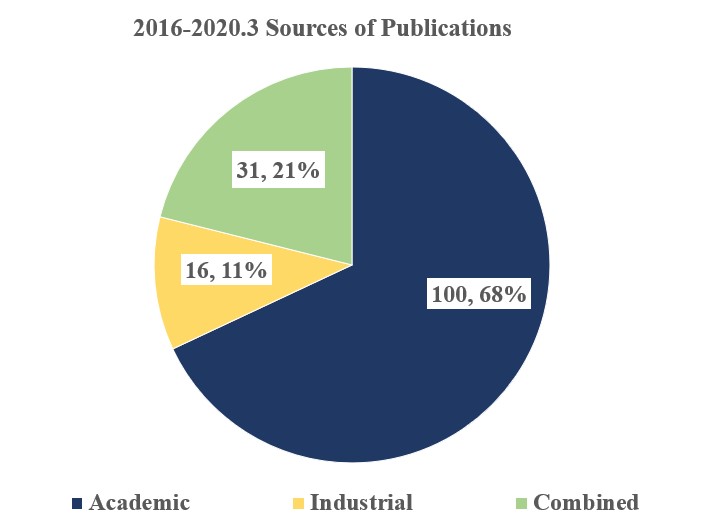}
	\caption{ (2016-2020.3) Sources of publications}
	\label{fig_3-3}
\end{figure}

The statistical results are shown in Fig. \ref{fig_3-3}. We counted 147 documents in total. From the figure, we can see that academic papers account for the largest proportion, reaching 68\%, followed by mixed-type 21\%, and finally industrial type 11\%. The statistical results indicate that academic institutions have the highest degree of participation. FL is in the preliminary stage of research, and various feasible technologies are stepping up experiments and research. The industry and mixed types also account for a part of the proportion. It shows that the industry also has a certain participation degree, and FL has been applied to some realistic scenes.

\par

\begin{figure}[htbp]
	\centering
	\includegraphics[width=3.4in]{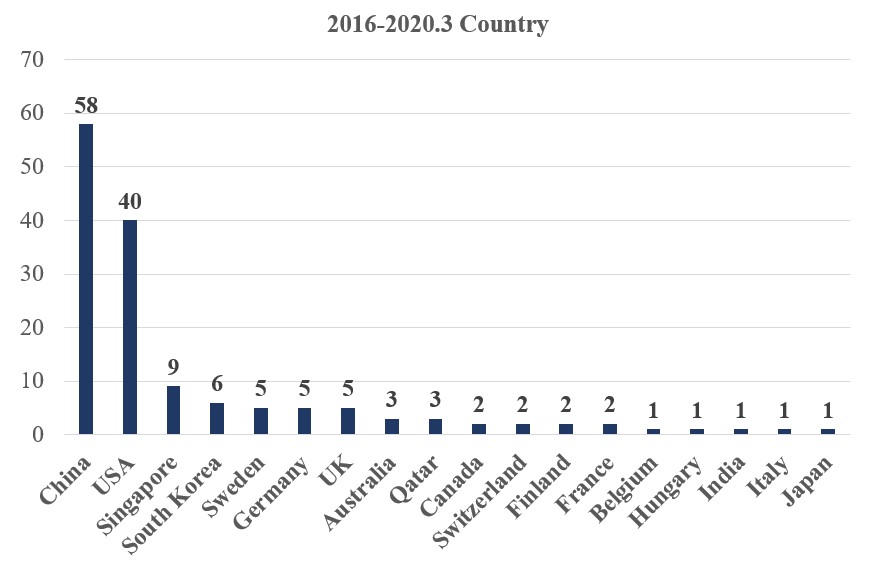}
	\caption{ (2016-2020.3) Distribution of countries of the first authors’ institutions}
	\label{fig_3-4}
\end{figure}

We counted the countries of the first author’s institutions of 147 articles. Fig. \ref{fig_3-4} shows the statistical result. China (39\%) and the United States (27\%) participated in the most FL researches, followed by Singapore, South Korea, Sweden, and Germany.

\subsubsection{RQ1.4 What are the application scenes of FL, and what practical problems are solved?}

We conducted a statistical analysis of the application scenes of FL. We referred to the article \cite{Y2019federatedLearning}. We divided the scenes into the following six categories:
\begin{itemize}[leftmargin=*]
	\item Edge Computing and Internet of Things
	\item Healthcare
	\item Urban Computing and Smart City
	\item Physical Information System
	\item Internet and Finance
	\item Industrial Manufacturing
\end{itemize}

\par
\begin{figure}[htbp]
	\centering
	\includegraphics[width=3.6in]{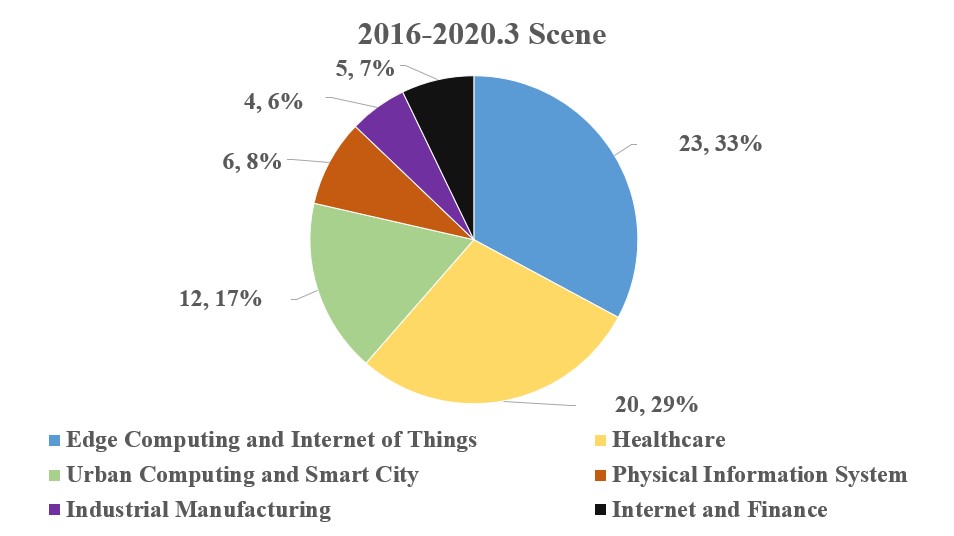}
	\caption{ (2016-2020.3) Application scenes of FL}
	\label{fig_3-5}
\end{figure}

Fig. \ref{fig_3-5} shows the statistical result. The three scenes of Edge Computing and Internet of Things (33\%), Healthcare (29\%), Urban Computing and Smart City (17\%) have the most applications. There are also a small number of applications on the Internet and Financial, physical Information System, and Industrial Scenes.
\par
The problems solved using FL are diverse, indicating that FL, like other AI techniques, can be universally applied to various fields. Nevertheless, FL is still rarely applied in specific areas, such as industrial manufacturing. There are serious data privacy protection problems in these scenarios and need to be solved by FL. Therefore, applying FL to solve practical problems in industrial manufacturing and other fields is still in urgent need of research.

\par

We further analyzed the statistics of the scenes, and conducted statistics for the specific problems solved in each scene. The statistical result is shown in Table \ref{table_3-1}.
\begin{table}[htbp]
	\small
	\caption{Scenes of FL}
	\label{table_3-1}
	\centering
	\begin{tabular}{p{2cm}<{\centering}|p{6cm}<{\raggedright}}
		\hline
		\textbf{Scene}  &  \textbf{Solved problem}   \\ \hline
		\multirow{20}{*}{\shortstack{Edge   \\Computing \\and Internet \\of Things}} & Network anomaly detection,   network intrusion detection \cite{DBLP:conf/soict/ZhaoCWTY19,DBLP:conf/icdcs/NguyenMMFAS19,DBLP:conf/bigdataconf/CetinLKSW19}, prediction of user association in the   network \cite{wang2020federated}\\ 
		& Virtual Reality \cite{DBLP:conf/globecom/ChenSSLY19}, Augmented   Reality \cite{chen2020federated}, AI object detection \cite{luo2019real} \\ 
		& Failure of electric vehicle   batteries and related accessories prediction \cite{DBLP:conf/hotedge/LuYS19} \\ 
		& Keyword spotting based on mobile   devices \cite{DBLP:conf/icassp/LeroyCLGD19}, mobile keyboard prediction \cite{yang2018applied,hard2018federated,DBLP:conf/conll/ChenSMWABR19}, mobile keyboard emoji   Prediction \cite{ramaswamy2019federated}, Out-Of-Vocabulary Words learning \cite{chen2019federated} \\ 
		& Human activity recognition based   on sensor data \cite{DBLP:conf/ispa/SozinovVG18}, EEG classification in human behavior and emotion based   on device data \cite{gao2019hhhfl}, Human mobility prediction based on mobile devices \cite{DBLP:journals/imwut/FengRSGL20} \\ 
		& Web quality of experience   prediction \cite{DBLP:conf/mobicom/IckinVF19}, Firefox search improvement \cite{hartmann2018federated} \\ 
		& Mobile device information   tracking \cite{bakopoulou2019federated}, edge device anomaly detection \cite{ito2020device} \\ 
		& Visual question answering \cite{DBLP:conf/aaai/LiuWGFZ20} \\ \hline
		\multirow{14}{*}{Healthcare} & Preterm-birth prediction \cite{DBLP:journals/corr/abs-1910-12191}, early gestational weight gain prediction \cite{DBLP:conf/sensys/PuriDKMSBMHLV19} \\ 
		& Make health decisions based on   clinical data \cite{DBLP:journals/artmed/LiTZZLDL20,DBLP:journals/jbi/HuangSQMDL19} \\ 
		& Health prediction   \cite{DBLP:journals/ijmi/BrisimiCMOPS18,choudhury2019differential,liu2018fadl,pfohl2019federated,cui2019federated,huang2018loadaboost} \\ 
		& Brain tumor segmentation   \cite{DBLP:conf/miccai/ShellerREMB18,DBLP:conf/miccai/LiMXRHZBCOCF19}, whole brain segmentation \cite{roy2019braintorrent}, functional magnetic resonance   imaging analysis \cite{li2020multi} \\ 
		& Drug related properties   prediction \cite{xiong2020facing}, collaborative drug discovery \cite{chen2020fl} \\ 
		& Pain expression recognition   \cite{tobis2019federated}, wearable healthcare \cite{chen2020fedhealth} \\ 
		& Medical named entity recognition   \cite{ge2020fedner} \\
		& Clinical Natural Language   Processing \cite{DBLP:conf/bionlp/LiuDM19} \\ \hline
		\multirow{11}{*}{\shortstack{Urban \\Computing \\and Smart \\City}} & The images collected by the car   camera classification and labeling \cite{DBLP:journals/access/YeYPH20} \\ 
		& Air quality assessment \cite{DBLP:conf/globecom/HuGLM18,DBLP:journals/access/Wang00G19} \\
		& Energy demand prediction for   electric vehicle network \cite{DBLP:conf/globecom/SaputraHNDMS19,mulya2020federated} \\ 
		& Traffic sign recognition \cite{albaseer2020exploiting} \\ 
		& Traffic flow prediction \cite{liu2020privacy} \\ 
		& Vehicle scheduling \cite{DBLP:journals/tcom/SamarakoonBSD20}, routing   \cite{samal2019time} \\
		& Indoor localization \cite{ciftler2020federated} \\ 
		& Security monitoring \cite{DBLP:conf/aaai/LiuHLHLCFCYY20} \\ 
		& Power demand response \cite{DBLP:conf/bigdataconf/0013SLY19} \\ \hline
		\multirow{5}{*}{\shortstack{Physical \\Information \\System}} & Place matching \cite{DBLP:conf/robio/LiWJ019} \\
		& Autonomous driving \cite{liu2019federated,liang2019federated} \\ 
		& Cloud robotic systems \cite{DBLP:journals/ral/LiuWL19} \\  
		& Drone jamming attack detection   \cite{DBLP:journals/access/MowlaTDC20} \\ 
		& Failure prediction in   aeronautics \cite{aussel2020combining} \\ \hline
		\multirow{5}{*}{\shortstack{Internet and\\ Finance}} & Credit card fraud detection \cite{DBLP:conf/bigdata2/YangZYL019} \\ 
		& Financial crimes detection \cite{suzumura2019towards} \\ 
		& Financial text recognition \cite{DBLP:conf/ictai/ZhuWHXX19} \\ 
		& Personalized search \cite{DBLP:conf/swarm/ChenSH19}\\ 
		& News recommendation \cite{qi2020fedrec} \\ \hline
		\multirow{4}{*}{\shortstack{Industrial \\Manufacturing}} & Industrial topic modeling \cite{DBLP:conf/cikm/JiangSTWZXY19}\\
		& Environment condition monitoring \cite{DBLP:conf/iwann/HuSCL19}\\ 
		& Product visual inspection \cite{DBLP:conf/iciar/HanYG19} \\ 
		& Sensor failure prediction \cite{basnayake2019federated} \\ \hline
	\end{tabular}
\end{table}

\subsection{RQ2: What methods to improve the quality of the FL learning algorithm and training method?}

\subsubsection{RQ2.1 Are the algorithms between clients the same? Are the client-side and server-side algorithm models the same? Are there constraints on the client-side and server-side algorithms?}

We conducted statistics on the categories of the client-side algorithms, server-side algorithms, and federated settings. The major types are the neural network, decision tree and random forest, logistic regression, support vector machine, and others. The federated setting refers to the category of the training process of FL (horizontal(H)/vertical(V)/transfer(V)/reinforcement(R)). The statistical results are shown in Table \ref{table_3-2}.

\begin{table*}[htbp]
	\caption{Statistics of client-side algorithms for federated learning}
	\label{table_3-2}
	\centering
	\begin{tabular}{m{2.7cm}<{\centering}|m{6.5cm}<{\centering}|m{1.2cm}<{\centering}|m{0.3cm}<{\centering}|m{5.2cm}<{\centering}}
		\hline
		\textbf{Algorithm type} &  \textbf{Client-side algorithm} &  \textbf{Federated setting} &  \textbf{No} &  \textbf{Article} \\ \hline
		\multirow{12}{*}{\shortstack{Neural \\Network}} & AE (Autoencoder) & H, T & 4 & \cite{DBLP:journals/jbi/HuangSQMDL19,DBLP:conf/bigdataconf/CetinLKSW19,ito2020device,liu2018secure} \\ \cline{2-5} 
		& ANN (Artificial Neural Network) & H & 2 & \cite{skatchkovsky2020federated,DBLP:conf/bionlp/LiuDM19}\\ \cline{2-5} 
		& CNN (Convolutional Neural Networks) & H, T, R & 35 & \cite{DBLP:conf/aistats/McMahanMRHA17,chen2019communication,DBLP:conf/iccd/DuanLCTRQL19,DBLP:journals/access/YeYPH20,chen2020focus,DBLP:conf/iclr/WangYSPK20,DBLP:conf/aiia/AnelliDNF19,das2019privacy,DBLP:conf/ictai/ZhuWHXX19,chen2020federated,DBLP:conf/itat/SzegediKH19,DBLP:conf/robio/LiWJ019,DBLP:conf/icassp/LeroyCLGD19,DBLP:conf/mod/SjobergGKJ19,DBLP:conf/bigdata2/YangZYL019,DBLP:conf/miccai/ShellerREMB18,DBLP:conf/iciar/HanYG19,DBLP:journals/entropy/XiaoCSV20,sattler2019clustered,tuor2020data,yoon2020federated,liu2019federated,DBLP:conf/aaai/LiuWGFZ20,tobis2019federated,chen2018federated,zhuo2019federated,ge2020fedner,qi2020fedrec,DBLP:conf/aaai/LiuHLHLCFCYY20,gao2019hhhfl,luo2019real,alotaibi2018wisdom,zeng2020fmore,jiao2020toward,DBLP:conf/icip/0003HWZS19} \\ \cline{2-5} 
		& DNN (Deep Neural Networks) & H & 8 & \cite{DBLP:conf/soict/ZhaoCWTY19,DBLP:journals/access/Wang00G19,DBLP:conf/globecom/SaputraHNDMS19,DBLP:conf/ispa/SozinovVG18,roy2019braintorrent,xiong2020facing,mulya2020federated,samal2019time} \\ \cline{2-5} 
		& ELM (Extreme Learning Machine) & H & 1 & \cite{ito2020device} \\ \cline{2-5} 
		& GAN (Generative Adversarial Networks) & H & 1 & \cite{DBLP:conf/ipps/HardyMS19} \\ \cline{2-5} 
		& MLP (Multi-layer Perceptron) & H, R & 13 & \cite{DBLP:conf/aistats/McMahanMRHA17,DBLP:conf/apnoms/KimH19a,DBLP:conf/middleware/NilssonSUGJ18,DBLP:conf/icml/YurochkinAGGHK19,DBLP:journals/access/MowlaTDC20,choudhury2019differential,liu2018fadl,ciftler2020federated,cui2019federated,zhuo2019federated,bui2019federated,li2020multi,corinzia2019variational}\\ \cline{2-5} 
		& NN (Neural Network) & H & 4 & \cite{chandiramani2019performance,DBLP:conf/mobicom/IckinVF19,chen2018federated,suzumura2019towards} \\ \cline{2-5} 
		& RNN (Recurrent Neural Network) & H, V & 24 & \cite{DBLP:conf/aistats/McMahanMRHA17,chen2019communication,DBLP:conf/iclr/WangYSPK20,DBLP:journals/corr/abs-1812-06127,DBLP:journals/corr/abs-1910-12191,DBLP:conf/globecom/HuGLM18,DBLP:conf/globecom/ChenSSLY19,DBLP:conf/hotedge/LuYS19,DBLP:conf/icdcs/NguyenMMFAS19,DBLP:conf/iwann/HuSCL19,yang2018applied,DBLP:journals/entropy/XiaoCSV20,sattler2019clustered,ramaswamy2019federated,hard2018federated,li2019federated,DBLP:conf/conll/ChenSMWABR19,chen2019federated,chen2018federated,bui2019federated,ge2020fedner,DBLP:journals/imwut/FengRSGL20,liu2020privacy,zeng2020fmore} \\ \hline
		\multirow{2}{*}{\shortstack{Decision Tree and \\Random Forest}} & DT (Decision Tree) & H, V & 2 & \cite{cheng2019secureboost,aussel2020combining} \\ \cline{2-5} 
		& RF (Random Forest) & H & 2 & \cite{DBLP:conf/mod/SjobergGKJ19,DBLP:journals/corr/abs-1812-06127}\\ \hline
		Logistic Regression & LR (Logistic Regression) & H, V, T & 8 & \cite{DBLP:conf/ispa/SozinovVG18,yang2019quasi,choudhury2019differential,pfohl2019federated,chen2018federated,agarwal2020federated,yang2019parallel,hardy2017private}\\ \hline
		Support Vector Machine & SVM (Support Vector Machine) & H & 5 & \cite{DBLP:journals/ijmi/BrisimiCMOPS18,DBLP:conf/swarm/ChenSH19,bakopoulou2019federated,choudhury2019differential,wang2020federated}\\ \hline
		Actor-critic & Actor-critic & R & 1 & \cite{DBLP:journals/sensors/LimKHH20} \\ \hline
		Maximum Likelihood Estimate & MLE (Maximum Likelihood Estimate) & H & 3 & \cite{DBLP:journals/tcom/SamarakoonBSD20,basnayake2019federated,DBLP:conf/sensys/PuriDKMSBMHLV19}\\ \hline
		\multirow{3}{*}{other} & MTRL (Multi-task Relationship Learning) & H & 1 & \cite{DBLP:conf/bigdataconf/LiMJ019} \\ \cline{2-5} 
		& QSAR (Quantitative structure-activity relationship) & H & 1 & \cite{chen2020fl}\\ \cline{2-5} 
		& Ranking algorithm & H & 1 & \cite{hartmann2018federated} \\ \hline
	\end{tabular}
\end{table*}

We collected 101 documents, and the algorithms between the client and client are the same. This result corresponds to the training process of federated learning. Client-side algorithms that can be used in horizontal federated learning are more common. Machine learning algorithms can be used for horizontal federated learning. Currently, only RNN (Recurrent Neural Network) \cite{DBLP:conf/iwann/HuSCL19}, DT (Decision Tree) \cite{cheng2019secureboost}, LR ( Logistic Regression) \cite{yang2019quasi,yang2019parallel,hardy2017private} are used for vertical federated learning; AE (Autoencoder) \cite{liu2018secure} CNN (Convolutional Neural Networks) \cite{liu2019federated}, LR (Logistic Regression) \cite{DBLP:conf/bigdataconf/Gao0HJYY19} are used for federated transfer learning; CNN (Convolutional Neural Networks) \cite{zhuo2019federated}, MLP (Multi-layer Perceptron) \cite{zhuo2019federated}, Actor-critic \cite{DBLP:journals/sensors/LimKHH20,liang2019federated} are used for federated reinforcement learning. In the future, vertical federated learning, federated transfer learning, and federated reinforcement learning still have many research spaces. How to apply common algorithms to these three federated settings still requires in-depth research.
\par
Horizontal federated learning aligns the feature space and expands the sample space. Each client has the same feature space and trains the same model, so neural network, machine learning, and some other algorithms can be used in horizontal federated learning; Vertical federated learning aligns the sample space and expands the feature space. Therefore, each client has different features of the same sample. Therefore, the server needs to consider how to aggregate each client’s model, which is more difficult; both federated transfer learning and federated reinforcement learning are generated by combining FL based on transfer learning and reinforcement learning. At present, these two types of federated scenes are relatively few, which results in the low use of these two types of federated settings.
\par
We conducted statistics on the server-side model and the client-side model of FL. A total of 88 articles were counted. The client-side model in 86 articles (98\%) was the same as the server-side model. There are two articles \cite{aussel2020combining,DBLP:conf/sensys/PuriDKMSBMHLV19}, in which the client-side model is different from the server-side model, as shown in Table \ref{table_3-3}. The models at both ends either have similar parameter structures, or there is a way to convert the client-side model to the server-side model. In the federated learning training process designed in \cite{aussel2020combining}, the client uses the Decision Tree (DT) algorithm. The server uses the Random Forest (RF). The server’s random forest is integrated from the client’s decision trees to predict the results. In the federated learning training process designed in \cite{DBLP:conf/sensys/PuriDKMSBMHLV19}, the client uses the Maximum Likelihood Estimate (MLE) algorithm, and the server uses the Maximum a posteriori estimation (MAP) algorithm, which is a combination of MLE and BE (Bayesian estimation). The coefficient obtained through MLE in the client is used as a priori estimate. The maximum likelihood estimation obtained from the client training data is combined with the global update’s prior distribution. The maximum coefficient of the fitted equation is calculated by the Bayesian formula posterior probability.

\begin{table*}[htbp]
	\caption{The difference between the client model and the server model}
	\label{table_3-3}
	\centering
	\begin{tabular}{m{6cm}<{\centering}|m{6cm}<{\centering}|m{4cm}<{\centering}|c}
		\hline
		\textbf{Model of Server} & \textbf{Model of Client} & \textbf{Federated Setting} & \textbf{Article} \\ \hline
		RF (Random Forest) & DT (Decision Tree) & H & \cite{aussel2020combining} \\ \hline
		MAP (Maximum a posteriori estimation) & MLE (Maximum Likelihood Estimate) & H & \cite{DBLP:conf/sensys/PuriDKMSBMHLV19} \\ \hline
	\end{tabular}
\end{table*}

\begin{table*}[htbp]
	\caption{Evaluation indicators detail}
	\label{table_3-4}
	\centering
	\begin{tabular}{m{5cm}<{\centering}|m{7cm}<{\centering}|m{5cm}<{\centering}}
		\hline
		\textbf{Name}	&\textbf{Meaning} &\textbf{Calculation method}\\  \hline
		ACC (Accuracy)	&Global accuracy& $ACC = \frac{{TP + TN}}{{TP + FN + TN + FP}}$\\  \hline
		AUC (Area Under Curve) AUC-ROC (Area Under ROC Curve)	&The area under the ROC curve, AUC generally refers to AUC-ROC & \tabincell{c}{$\frac{{\sum {I{P_{pos}}{P_{neg}}} }}{{M*N}}$
			\\
			$M$:Number of positive samples\\$N$:Number of negative samples
		}\\
		\hline
		RMSE (Root Mean Squared Error)	&Square root of the ratio of sum of squared deviations between observations and true values to number of observations n& $RMSE = \sqrt {\frac{1}{n}{{\sum\limits_{i = 1}^n {({y_i} - \widehat {{y_i}})} }^2}} $
		\\  \hline
	\end{tabular}
\end{table*}

\par
At present, most articles still adopt a unified client and server model or use different models such as \cite{aussel2020combining,DBLP:conf/sensys/PuriDKMSBMHLV19}, but they are essentially the same type. The model training process of the FL client has a certain degree of independence. Can different clients choose different learning algorithms suitable for themselves so that the client training process has stronger autonomy? In the future, the integration method of different models of the client can be studied.

\subsubsection{RQ2.2 Considering the client-side, what are the methods to improve the quality of FL?}
The evaluation indicators are explained in Table \ref{table_3-4}. In order to better explain the evaluation indicators, here, we explain TP, TN, FP, FN. T represents the prediction is equal to the actual; F represents the prediction is not equal to the actual; P represents the prediction result is a positive example; N represents the prediction result is a negative example. TP means that the prediction is positive and the actual is positive; FP means that the forecast is positive and the actual is negative; TN means that the forecast is negative, and the actual is negative; FN means that the forecast is anti and the actual is positive. TP+TN represents the correct number of predictions.

We focused on model quality. We conducted statistical analysis from the improvement methods and effects of the client algorithm model, as shown in Table \ref{table_3-5}.

\begin{table*}[htbp]
	\small
	\caption{Client-side improvement algorithm}
	\label{table_3-5}
	\centering
	\begin{tabular}{m{6.3cm}<{\centering}|m{1.5cm}<{\centering}|m{2.7cm}<{\centering}|m{2cm}<{\centering}|m{2cm}<{\centering}|c}
		\hline
		\multirow{2}{*}{\textbf{Improvement method}} & \multicolumn{4}{c|}{\textbf{Improvement effect}} & \multirow{2}{*}{\textbf{Article}} \\ \cline{2-5}
		& \textbf{Indicator} & \textbf{Proposed method} & \textbf{Other method} & \textbf{Improvement (\%)} &  \\ \hline
		Federated Multi-view Learning (fedMVL) & ACC & 0.8295 & 0.7094 & 12.01 & \cite{huang2019Iterative} \\ \hline
		\multirow{2}{*}{\shortstack{Loss-based Adaboost Federated Machine \\Learning (LoAdaBoost)}} & AUC & 0.7931 & 0.7883 & 0.48 & \multirow{2}{*}{\cite{huang2018loadaboost}} \\ \cline{2-5}
		& \multicolumn{4}{c|}{Reduced communication cost} &  \\ \hline
		Federated Multi-task Learning & ACC & 0.56 & 0.45 & 11 & \cite{corinzia2019variational} \\ \hline
		Federated personalization & ACC & \multicolumn{3}{c|}{1\% higher than non-FL training} & \cite{tobis2019federated} \\ \hline
		Federated Meta-learning (FedMeta) & ACC & 0.8094 & 0.7586 & 5.08 & \cite{chen2018federated} \\ \hline
		Federated User Representation Learning (FURL) & AUC & 0.6241 & 0.119 & 50.51 & \cite{bui2019federated} \\ \hline
		Federated Continuous Learning (FCL) & ACC & 0.8471 & 0.7955 & 5.16 & \cite{yoon2020federated} \\ \hline
	\end{tabular}
\end{table*}

Most articles proposed new FL algorithms based on existing machine/deep learning algorithms from the statistical results.

\begin{itemize}[leftmargin=*]
	\item Federated Multi-view Learning (FedMVL) \cite{huang2019Iterative}, multi-view learning aims to build a global model using shared views collected from multiple independent sources. Federated multi-view learning, combining FL with multi-view learning, can improve the performance and effect of the view learning model while solving communication and computing efficiency problems.
	\item Loss-based Adaboost Federated Machine Learning (LoAdaBoost) \cite{huang2018loadaboost}, which combines Adaboost with FL. Adaboost is an iterative algorithm whose core idea is to train different classifiers (weak classifiers) for the same training set. Then these weak classifiers are combined to form a stronger final classifier (strong classifier). This method can improve the learning effect of FL.
	\item Federated Multi-task Learning \cite{corinzia2019variational}. Multi-task learning is an effective method on real data sets. It aims to use other related tasks to improve the generalization ability of the main tasks. The effect of multi-task FL is better than traditional FL.
	\item Federated personalization \cite{tobis2019federated}, limits the number of shared model parameters in a deterministic way, and only allows a fixed subset of model parameters to be shared with the central server, which adds a layer of privacy protection to the FL algorithm.
	\item Federated Meta-learning (FedMeta) \cite{chen2018federated}. The goal of meta-learning is to meta-train an algorithm that can quickly train models, such as deep neural networks, to complete new tasks. The goal of federated meta-learning is to use data distributed between clients to train meta-algorithms collaboratively.
	\item Federated User Representation Learning (FURL) \cite{bui2019federated}. User representation learning is a personalized method that can significantly improve prediction accuracy based on neural network models. Combining FL with user representation learning can improve the quality of the model.
	\item Federated Continuous Learning (FCL) \cite{yoon2020federated}. The purpose of continuous learning or lifelong learning is to prevent catastrophic forgetting. When the neural network learns new tasks, the knowledge of the old tasks is not covered. It combines continuous learning with FL to continuously learn the local model on each client while allowing it to use indirect experience (task knowledge) from other clients to improve the model’s quality. Besides, this method can reduce the amount of communication data.
\end{itemize}

\par
Among them, the most apparent improvement is federated user representation learning \cite{bui2019federated}, federated multi-view learning \cite{huang2019Iterative}, and federated multi-task learning \cite{corinzia2019variational}.

\subsubsection{RQ2.3 Considering the server-side, what are the aggregation algorithms to improve the quality of FL?}

Server-side model/parameter aggregation is an important part of FL. Many researchers have studied the server-side aggregation algorithm of FL. We counted the aggregation algorithms used in FL. We conducted statistics on federated settings, design (improvement) goals, types, and client factors considered. Through statistics, there are two types of aggregation for FL:

\begin{itemize}[leftmargin=*]
	\item Parameter aggregation (P): Aggregate according to parameters or intermediate calculation parameters of the client-side models.
	\item Model aggregation (M): Aggregate according to the client-side models.
\end{itemize}

\par
Through statistics, the design goals of the aggregation algorithm of FL are often considered from four aspects: (1) From the aspect of algorithm construction, design the basic aggregation algorithm according to the training process of FL; (2) From the aspect of model accuracy, improve the quality of training models; (3) From the aspect of computational efficiency, accelerate model convergence; (4) From the aspect of communication efficiency, reduce the amount of communication data during the training process.
\par
The statistical results are shown in Table \ref{table_3-6}.

\begin{table*}[htbp]
	\small
	\caption{Statistics of basic aggregation algorithms for federated learning}
	\label{table_3-6}
	\centering
	\begin{tabular}{m{2cm}<{\centering}|m{1.2cm}<{\centering}|m{2cm}<{\centering}|c|m{4.5cm}<{\centering}|c|m{4cm}<{\centering}}
		\hline
		\textbf{Aggregation algorithm} & \textbf{FL Setting} & \textbf{Design (improvement) goals} & \textbf{Type} & \textbf{Considerations of clients} & \textbf{No} & \textbf{Article} \\ \hline
		FedSGD & H, R & \multirow{7}{*}{\shortstack{Basic \\aggregation \\algorithm}} & P & Gradient, Quantity of data & 10 & \cite{DBLP:conf/aistats/McMahanMRHA17,DBLP:conf/iccd/DuanLCTRQL19,DBLP:conf/icassp/LeroyCLGD19,DBLP:conf/ipps/HardyMS19,chen2018federated,DBLP:journals/sensors/LimKHH20,agarwal2020federated,ge2020fedner,qi2020fedrec,samal2019time}\\ \cline{1-2} \cline{4-7} 
		RF & \multirow{5}{*}{H} &  & M & Model & 2 & \cite{DBLP:journals/artmed/LiTZZLDL20,aussel2020combining} \\ \cline{1-1} \cline{4-7} 
		ILTM &  &  & P & Model similarity, Training parameters & 1 & \cite{DBLP:conf/cikm/JiangSTWZXY19} \\ \cline{1-1} \cline{4-7} 
		MAP &  &  & P & Prior distribution & 1 & \cite{DBLP:conf/sensys/PuriDKMSBMHLV19} \\ \cline{1-1} \cline{4-7} 
		DT &  &  & P & Parameters of decision tree & 1 & \cite{cheng2019secureboost} \\ \cline{1-1} \cline{4-7} 
		PFNM &  &  & M & Model & 1 & \cite{DBLP:conf/icml/YurochkinAGGHK19} \\ \cline{1-2} \cline{4-7} 
		QNA & V &  & P & Curvature & 1 & \cite{yang2019quasi} \\ \hline
		FOCUS & \multirow{7}{*}{H} & \multirow{7}{*}{\shortstack{Improve \\model \\quality \\(improvement)}} & M & Model, Quality of model & 1 & \cite{chen2020focus}\\ \cline{1-1} \cline{4-7} 
		FedMA &  &  & P & Weight of each network layer & 1 & \cite{DBLP:conf/iclr/WangYSPK20}\\ \cline{1-1} \cline{4-7} 
		FedProx &  &  & M & Model, Quantity of data & 1 & \cite{DBLP:journals/corr/abs-1812-06127} \\ \cline{1-1} \cline{4-7} 
		FUALA &  &  & M & Model, Quality of model & 1 & \cite{DBLP:journals/corr/abs-1910-12191} \\ \cline{1-1} \cline{4-7} 
		PMA &  &  & M & Quantity of data, Diversity of label, Differences of model & 1 & \cite{DBLP:conf/aiia/AnelliDNF19} \\ \cline{1-1} \cline{4-7} 
		OFMTL &  &  & M & Model, Quality of model & 1 & \cite{DBLP:conf/bigdataconf/LiMJ019}\\ \cline{1-1} \cline{4-7} 
		SCAFFOLD &  &  & M & Model & 1 & \cite{DBLP:journals/corr/abs-1910-06378} \\ \hline
		FedAvg & H, T, R & Converge faster (improvement) & M & Model, Quantity of data & 58 &  \cite{DBLP:conf/aistats/McMahanMRHA17,DBLP:conf/middleware/NilssonSUGJ18,DBLP:conf/soict/ZhaoCWTY19,das2019privacy,DBLP:conf/bigdataconf/Gao0HJYY19,DBLP:conf/ictai/ZhuWHXX19,DBLP:conf/globecom/HuGLM18,DBLP:conf/globecom/ChenSSLY19,DBLP:journals/access/Wang00G19,chen2020federated,DBLP:journals/jbi/HuangSQMDL19,DBLP:conf/hotedge/LuYS19,DBLP:conf/icdcs/NguyenMMFAS19,DBLP:journals/tcom/SamarakoonBSD20,DBLP:conf/globecom/SaputraHNDMS19,DBLP:conf/swarm/ChenSH19,DBLP:conf/mod/SjobergGKJ19,DBLP:journals/access/MowlaTDC20,DBLP:conf/bigdataconf/CetinLKSW19,DBLP:conf/bigdata2/YangZYL019,DBLP:conf/ispa/SozinovVG18,DBLP:conf/miccai/ShellerREMB18,DBLP:conf/mobicom/IckinVF19,bakopoulou2019federated,yang2018applied,DBLP:journals/entropy/XiaoCSV20,tuor2020data,choudhury2019differential,albaseer2020exploiting,xiong2020facing,liu2018fadl,pfohl2019federated,hartmann2018federated,ramaswamy2019federated,basnayake2019federated,ciftler2020federated,hard2018federated,li2019federated,DBLP:conf/conll/ChenSMWABR19,chen2019federated,cui2019federated,tobis2019federated,skatchkovsky2020federated,liang2019federated,bui2019federated,DBLP:conf/aaai/LiuHLHLCFCYY20,chen2020fl,gao2019hhhfl,huang2018loadaboost,li2020multi,DBLP:journals/imwut/FengRSGL20,liu2020privacy,luo2019real,DBLP:conf/bionlp/LiuDM19,alotaibi2018wisdom,zeng2020fmore,jiao2020toward,DBLP:conf/icip/0003HWZS19}\\ \hline
		SMA & \multirow{7}{*}{H} & \multirow{8}{*}{\shortstack{Converge faster \\and improve \\model quality \\(improvement)}} & M & Data quality, Computing power, type & 1 & \cite{DBLP:journals/access/YeYPH20}\\ \cline{1-1} \cline{4-7} 
		NDW &  &  & M & Model, Quantity of data and model, Frequency of participation & 1 & \cite{DBLP:conf/apnoms/KimH19a} \\ \cline{1-1} \cline{4-7} 
		ASTW &  &  & M & Model, Quantity of data, Sequence of recently updated models & 1 & \cite{chen2019communication} \\ \hline
		FNE & H & Reduce communication (improvement) & P & Quantity of data points, Model's abstract fitness number & 1 & \cite{DBLP:conf/itat/SzegediKH19} \\ \hline
	\end{tabular}
\end{table*}

\subsubsection{RQ2.4 Considering the server-side, what methods improve the aggregation algorithm to enhance the training quality?}
We give corresponding definitions and explanations for some nouns that appear later. The FedAvg algorithm and FedSGD algorithm are shown in Table \ref{table_3-7}. Both FedAvg and FedSGD algorithms are given in the article \cite{DBLP:conf/aistats/McMahanMRHA17}. The idea of the FedSGD algorithm is to update the server-side model using the traditional gradient update method, while the FedAvg algorithm selects some clients each time and updates the server model through weighted average client models. FedAvg algorithm can converge faster. FedAvg algorithm reduces the number of communication rounds required for convergence by 90\% compared to FedSGD \cite{DBLP:conf/aistats/McMahanMRHA17}.

\begin{table*}[htbp]
	\caption{Description of aggregation algorithms FedAvg, FedSGD}
	\label{table_3-7}
	\centering
	\begin{tabular}{c|c|c}
		\hline
		\textbf{Method}	&\textbf{Full name} &\textbf{Aggregation process of server}\\
		\hline
		FedAvg	&Federated Averaging &\tabincell{c}{${w_{t + 1}} \leftarrow \sum\nolimits_{k = 1}^K {\frac{{{n_k}}}{n}} w_{t + 1}^k$\\
			$w_{t + 1}^k$:The model of the kth client}\\  \hline
		FedSGD	&\tabincell{c}{Federated Stochastic Gradient Descent}&\tabincell{c}{
			${w_{t + 1}} \leftarrow {w_t} - \eta \sum\nolimits_{k = 1}^K {\frac{{{n_k}}}{n}} {g_k}$\\
			${g_k}$:The gradient of the kth client}\\  \hline
	\end{tabular}
\end{table*}

\par
We conducted a statistical analysis of the improvement goals and effects of the aggregation algorithms. Since the FedAvg algorithm is currently the most used in FL, we compared the FedAvg algorithm as a baseline with other aggregation algorithms. The statistical result is shown in Table \ref{table_3-8}.

\begin{table*}[htbp]
	\caption{Statistics of aggregation algorithms for federated learning}
	\label{table_3-8}
	\centering
	\begin{tabular}{m{2cm}<{\centering}|m{3.5cm}<{\centering}|c|m{4cm}<{\centering}|c|c}
		\hline
		\multirow{3}{*}{\shortstack{\textbf{Aggregation }\\\textbf{ algorithm}}} & \multirow{2}{*}{\textbf{Improvement goal}} & \multicolumn{4}{c}{\textbf{Effect}} \\ \cline{3-6} 
		&  & \textbf{Indicator} & \textbf{Proposed method} & \textbf{FedAvg} & \textbf{Improvement(\%)} \\ \hline
		FOCUS & \multirow{5}{*}{Improve model quality} & ACC & 0.7008 & 0.6426 & 5.82 \\ \cline{1-1} \cline{3-6} 
		FedMA &  & ACC & 0.4907 & 0.4663 & 2.44 \\ \cline{1-1} \cline{3-6} 
		FedProx &  & ACC & 0.75 & 0.68 & 7 \\ \cline{1-1} \cline{3-6} 
		FUALA &  & AUC & 0.678 & 0.622 & 5.6 \\ \cline{1-1} \cline{3-6} 
		SCAFFOLD &  & ACC & 0.842 & 0.828 & 1.4 \\ \hline
		\multirow{2}{*}{SMA} & \multirow{6}{*}{\shortstack{Converge faster and \\improve model quality}} & ACC & 0.72 & 0.65 & 7 \\ \cline{3-6} 
		&  & \multicolumn{4}{c}{Converges faster than FedAvg} \\ \cline{1-1} \cline{3-6} 
		\multirow{2}{*}{NDW} &  & ACC & 0.87 & 0.82 & 5 \\ \cline{3-6} 
		&  & \multicolumn{4}{c}{Converges faster than FedAvg} \\ \cline{1-1} \cline{3-6} 
		\multirow{2}{*}{ASTW} &  & ACC & 0.9712 & 0.95 & 2.12 \\ \cline{3-6} 
		&  & \multicolumn{4}{c}{The number of communication rounds is reduced by 50\% compared to FedAvg} \\ \hline
		FNE & Reduce communication & \multicolumn{4}{c}{Reduce the amount of data during communication} \\ \hline
	\end{tabular}
\end{table*}

\begin{table*}[htbp]
	\small
	\caption{Federated learning improved algorithm statistics}
	\label{table_3-9}
	\centering
	\begin{tabular}{m{1.3cm}<{\centering}|m{2.8cm}<{\centering}|m{4.5cm}<{\centering}|m{1.5cm}<{\centering}|m{4.5cm}<{\centering}|c}
		\hline
		\multirow{2}{*}{\textbf{Purpose}} & \multirow{2}{*}{\textbf{Improve method}} & \multirow{2}{*}{\textbf{Specific description}} & \multicolumn{2}{c|}{\textbf{Effect}} & \multirow{2}{*}{\textbf{article}} \\ \cline{4-5}
		&  &  & \textbf{Indicator} & \textbf{Specific description} &  \\ \hline
		\multirow{23}{*}{\shortstack{Improve \\model \\quality}} & \multirow{9}{*}{\shortstack{Add an algorithm \\based on client \\clustering}} & Propose Federated Region-learning, divide monitoring points into regions, and aggregate according to regions & ACC & 0.82\% higher than centralized learning & \cite{DBLP:conf/globecom/HuGLM18} \\ \cline{3-6} 
		&  & Propose Community-based Federated Machine Learning (CBFL), add cluster processing & AUC & 9.1\% higher than traditional FL & \cite{DBLP:journals/jbi/HuangSQMDL19} \\ \cline{3-6} 
		&  & Add the client clustering process & RMSE & Reduced from 5.81 to 5.76 & \cite{DBLP:conf/globecom/SaputraHNDMS19,mulya2020federated} \\ \cline{3-6} 
		&  & Propose Clustered Federated Learning (CFL) & ACC & When the data is a cluster structure, it is 33\% higher than traditional FL & \cite{sattler2019clustered} \\ \cline{2-6} 
		& Optimization model initialization method & Propose Federated Feature Fusion (FedFusion) & ACC & 2.85\% higher than FedAvg, and the number of communication rounds is reduced by 60\% & \cite{DBLP:conf/icip/0003HWZS19} \\ \cline{2-6} 
		& \multirow{6}{*}{\shortstack{Solve the problem \\of unbalanced data\\ distribution}} & Data expansion based on global data distribution and multi-client rescheduling based on intermediary to improve accuracy & ACC & Improve by 5\% on unbalanced data sets & \cite{DBLP:conf/iccd/DuanLCTRQL19} \\ \cline{3-6} 
		&  & Propose Federated-autonomous Deep Learning (FADL) & AUC & 4\% higher than traditional FL, and reach the level of centralized learning & \cite{liu2018fadl} \\ \cline{2-6} 
		& Select the clients & Based on the benchmark data set, evaluate and select the appropriate clients for training & ACC & Achieve an effect level without noise & \cite{tuor2020data} \\ \hline
		Avoid single points of failure & Decentralization & Change the server model in FL to point-to-point model & ACC & 1.5\% higher than FedAvg & \cite{roy2019braintorrent} \\ \hline
	\end{tabular}
\end{table*}

We discussed and analyzed the design (improvement) goals of the aggregation algorithms. The client factors considered the application scenes and effects.
\par

From the perspective of design goal, FedSGD, PFNM, RF, ILTM, MAP, DT are designed from the aspect of algorithm construction. The FedAvg and ASTW algorithms improve the convergence speed. The FOCUS, FedMA, FedProx, FUALA, PMA, and OFMTL algorithms improve model accuracy. The SMA and NDW algorithms improve both the convergence speed and the model accuracy. The SCAFFOLD algorithm solves the problem of training bias caused by data heterogeneity. The FNE algorithm reduces the amount of data of the clients in the aggregation.
\par

From the perspective of the client factors considered, in addition to the traditional parameters considered, NDW also considers the amount of data, model accuracy, client participation frequency. SMA, FOCUS targets the data quality of the client. PMA considers the label diversity, model differences; ILTM targets the differences between models.
\par

From the perspective of application scenes, SMA, FedProx, and SCAFFOLD are used when the client data is heterogeneous; FOCUS and FUALA are used when the client data quality is uneven.
\par

From the perspective of effects, the ASTW, FOCUS, FedMA, FedProx, FUALA, SCAFFOLD, SMA, NDW algorithms are better than FedAvg. Among them, the FedProx, SMA, and FOCUS three algorithms have the most remarkable improvement.
\par

FedProx \cite{DBLP:journals/corr/abs-1812-06127} shows a more stable and accurate convergence behavior compared to FedAvg in a highly heterogeneous environment. Based on FedAvg, FedProx allows clients with available system resources to perform a variable amount of work locally and then aggregates partial solutions for some clients instead of discarding them.
\par

SMA \cite{DBLP:journals/access/YeYPH20} (Selective Model Aggregation) is a selective model aggregation method, which is superior to the FedAvg method in terms of accuracy and efficiency. The method of model selection is described as a two-dimensional image calculation process-theoretical issue, through screening models with a better quality of raw data and models.
\par

FOCUS \cite{chen2020focus} calculates the difference between the FL model’s performance on the local dataset and the client’s local FL model on the benchmark dataset to reduce the impact of the client’s label noise on the global model training. The method calculates cross-entropy to quantify the credibility of the client-side data without observing the client-side data. The method can effectively identify clients with noisy labels and reduce their impact on model performance, thereby improving the global model’s quality.

\subsubsection{RQ2.5 Considering the server-side, what methods to improve the training quality by enhancing the federated training process?}
We focused on model quality issues, but some articles focused on model quality with some other improvements, such as avoiding single points of failure, etc. We made statistics on these improved algorithms according to the improvement purpose, improvement methods, and improvement effect. The statistical results are shown in Table \ref{table_3-9}.

It can be seen from the table that the process of improving FL from the server side is mainly considered from the following five aspects:
\par

\textbf{(1) Based on the client clustering algorithm, improve the quality of the FL model.} The articles \cite{DBLP:conf/globecom/HuGLM18,DBLP:journals/jbi/HuangSQMDL19,DBLP:conf/globecom/SaputraHNDMS19,xiong2020facing,mulya2020federated} added clustering algorithm, which can divide clients, and aggregate models of the same type of clients, which improves the quality of the model trained after clustering.
\par

\textbf{(2) Optimization model initialization method.} The article \cite{DBLP:conf/icip/0003HWZS19} added feature fusion to the process of FL. Feature fusion combines the features from local and global models in an efficient and personalized way. This method provides better initialization for new clients.
\par

\textbf{(3) Solve the problem of unbalanced data distribution.} The article \cite{DBLP:conf/iccd/DuanLCTRQL19} used data expansion based on global data distribution and multi-client rescheduling based on a mediator to solve the problem of data imbalance. The article \cite{liu2018fadl} proposed a new Federated Autonomous Learning (FADL) method that balances global and local training. This method uses data from all data sources to train the first half of the neural network globally and then trains the second half of the neural network locally to deal specifically with each data source.
\par

\textbf{(4) Choose clients.} The article \cite{tuor2020data} evaluated the correlation of a single data sample on each client based on a benchmark model trained on a small benchmark data set for a specific task, then selected data with sufficiently high correlation. Only a selected subset of each client’s data is used in the collaborative learning process.
\par

\textbf{(5) Decentralization.} The article \cite{roy2019braintorrent} decentralized the process of model aggregation to each client, which can improve the reliability of the training process.
\par

The most obvious improvement is the algorithm based on client clustering \cite{sattler2019clustered}.

\subsection{RQ3: What methods to improve the data privacy protection of FL?}

We analyzed a total of 20 articles related to the privacy protection technology of FL. After statistical analysis, we can improve the data privacy of FL from two aspects: (1) Use encryption algorithms in the process of data transmission and training; (2) By improving the training process, weaken the connection between the model and the original data to improve privacy.
\par

\textbf{(1) Use encryption algorithm during data transmission and training}
\par
The current articles used five methods to ensure privacy, including differential privacy, homomorphic encryption, secure multi-party computing, and a trusted execution environment. Besides, there are hybrid methods that combine multiple techniques.
\par

\textbf{Differential Privacy.}
Articles \cite{wei2020performance,geyer2017differentially,DBLP:journals/tii/LuHDMZ20,DBLP:conf/miccai/LiMXRHZBCOCF19} used a differential privacy method to add noise to the client data to hide the real information of clients. This method requires a trade-off between the level of privacy protection and model performance. \cite{wei2020performance,geyer2017differentially} pointed out that the data privacy of the client can be protected at a small cost in terms of model performance.
\par

Articles \cite{bhowmick2018protection,DBLP:journals/tii/HaoLLXYL20} designed privacy protection mechanisms for FL based on differential privacy. Article \cite{bhowmick2018protection} further proposed an optimal local differential privacy mechanism on the basis of differential privacy. The author verified through experiments that this method is suitable for large-scale image classification and language models without reducing its practicability. Article \cite{DBLP:journals/tii/HaoLLXYL20} proposes a Privacy-enhanced FL (PEFL) scheme on the basis of differential privacy, which is non-interactive in each aggregation. Even if the opponent colludes with multiple honest entities, it also provides a high level of privacy protection.
\par

\textbf{Homomorphic Encryption.}
Article \cite{DBLP:conf/globecom/ZhangCYD19} used homomorphic encryption technology to achieve a high level of accuracy while maintaining a low computational cost.
\par

Article \cite{DBLP:conf/bigdataconf/FengHXSYZWCYL19} used Partial Homomorphic Encryption (PHE) and multi-party computing protocols, and uses a set of Partial Homomorphic Encryption (PHE) computing models and multi-party computing protocols to cover the key operators of distributed LightGBM learning and inference on both sides. The use of PHE and multi-party calculation model causes the congestion of calculation and communication overhead. While ensuring the statistical accuracy of the learning model, a certain statistical acceleration strategy is proposed to reduce the communication cost.
\par

\textbf{Multi-party Secure Computing based on encryption.}
Article \cite{zhu2020privacy-preserving} defined encrypted FL as a weighted FL problem and designed an oracle-assisted secure multi-party computing framework.
\par

\textbf{Trusted Execution Environment based on encryption.}
Articles \cite{MoEfficient2019,DBLP:journals/isci/ChenLLXLL20} used trusted execution environment to ensure the privacy of FL. Article \cite{MoEfficient2019} pointed out that the trusted execution environment is used to protect the privacy of machine learning. By allocating private areas of computing resources (such as memory), it can provide hardware and software isolation. Compared with traditional software protection such as homomorphic encryption, Trusted Execution Environment provides lower overhead and higher privacy. Article \cite{DBLP:journals/isci/ChenLLXLL20} evaluated the performance of trusted execution environment schemes through prototype implementation methods. The results show that the program has the characteristics of complete and practical training, and at the same time, it can ensure that the training process can be performed in a safe learning environment.
\par

\textbf{Hybrid method.}
Articles \cite{DBLP:conf/ccs/XuBZAL19,mugunthan2020privacyfl} used both differential privacy and multi-party secure computing technology. The method proposed in \cite{DBLP:conf/ccs/XuBZAL19} can reduce training time by 68\% and data transmission by 92\% without sacrificing privacy protection or model performance. By complying with the differential privacy framework, it is guaranteed that data cannot be reversed from intermediate results. Multi-party secure calculations ensure that information will not be leaked, thus providing participants with end-to-end privacy guarantees. Article \cite{mugunthan2020privacyfl} built an extensible, easy to configure, and extensible FL environment simulator based on differential privacy and secure multi-party computing methods.
\par

In addition, the article \cite{choudhury2020anonymizing} proposed a syntactic anonymization approach, which applies syntactic anonymization to the original private data and uses the generated anonymous data for subsequent mining. This method operates on data records composed of relational parts and transactional parts, thereby providing protection against adversaries from obtaining personal information that spans these two data types. There are also privacy protection technologies for specific machine learning methods. Article \cite{DBLP:conf/ccs/MandalG19} designed a multi-party regression training privacy protection protocol for the logistic regression model. The accuracy of this method is very close to that of no privacy protection measures. Article \cite{Feng2020Practical} proposed a privacy protection method for deep neural networks. This method can well support the nonlinear activation function in the encrypted domain so as to effectively train the deep neural network on the basis of protecting the privacy of the server-side model. At the same time, this method can be combined with secret sharing technology to further protect the privacy of client training data.
\par

The encryption algorithms currently used for FL are all existing, relatively mature algorithms. Many of them are encryption algorithms used in distributed computing. Most algorithms have been theoretically proven to achieve data privacy protection. Among the 16 articles, only \cite{DBLP:conf/bigdataconf/FengHXSYZWCYL19} is a vertical federated setting, the others are all horizontal federated settings, only \cite{DBLP:conf/ccs/MandalG19,Feng2020Practical} are for specific machine learning methods, and the rest of the articles are for privacy protection for the process of FL neural network model training. The differential privacy method will lose the accuracy of the model and easily lead to lower accuracy when the amount of data on each client is small; homomorphic encryption cannot satisfactorily solve the problem of nonlinear activation function in the encryption domain. The cost of calculation and communication are higher; multi-party secure calculation methods have higher calculation and communication costs for deep neural networks. In the future, suitable encryption methods should be proposed for other federated settings, such as vertical FL, federated transfer learning, and federated reinforcement learning.
\par

\textbf{(2) Improve the training process, weaken the connection between the model and the original data to improve privacy}
\par

Articles \cite{tobis2019federated,chen2018federated,bui2019federated} improved privacy from the client-side. Federated personalization \cite{tobis2019federated} determines the parameters shared with the server based on the location of the client. This method ensures that the server cannot access all parameters in the neural network, which improves privacy. Federated Meta-learning (FedMeta) \cite{chen2018federated} only transmits parameterized algorithms to ensure data privacy. Federated User Representation Learning (FURL) \cite{bui2019federated} divides model parameters into federated parameters and private parameters. Private parameters are trained locally and will not be transmitted to the server, thereby improving privacy.
\par

Article \cite{cui2019federated} improved privacy from the server-side. The author used a hybridization algorithm to eliminate the connection between the original data and model parameters, thereby avoiding data leakage. While improving privacy, the AUC of the proposed method is higher than the traditional FL.
\par

This method is unique for FL in the privacy protection mechanism. This RQ is open. The current research results are small in quantity and new in content. There are still many problems in this direction that can be studied.

\subsection{RQ4: What are the incentive mechanisms to improve the data quality of FL?}

In the process of FL, the server hopes to attract more clients with high-quality data to participate in the training at the minimum or controllable cost. The server-side incentive mechanism should incentivize the client to provide high-quality data to participate in the training phase, not just consider paying the price. Also, clients will misrepresent information in order to obtain more payment returns. Therefore, the incentive mechanism also needs to measure the credibility of the participants. This shows that a reasonable incentive mechanism needs to guarantee the level of participation and the quality of completion. 
\par

The meaning of the incentive mechanism is the organic synthesis of incentives and mechanisms. Mechanism refers to the form of the interaction, mutual connection, mutual restraint between the various elements of the system, and the principle of dynamic change. The incentive mechanism for FL is: the incentive subject system (server) uses one or more incentive methods to standardize management and interacts and restricts each other with the incentive object (client) through incentive factors. 
\par

By clarifying the application scenes of the incentive mechanism, we further determine the optimization goal of the incentive mechanism and then consider the system elements that should be included, and then use the system elements to design the corresponding incentive function, and finally determine the interaction mechanism and return method. The six aspects that should be considered in the design of the incentive mechanism are as follows:
\par

\textbf{1. Scene:} The scene where the incentive mechanism functions, that is, the scene in which the incentive mechanism is used.
\par

\textbf{2. Optimization goals (OG):} Where are the issues that the incentive mechanism cares about, and which optimization goals are the focus of the incentive mechanism. Among them, the optimization goals are divided into the following categories:
\par

\textbf{1) Improve model quality (Q).} Refers to an incentive mechanism based on users (type, contribution, reputation). User type refers to the type of data held by the user. User contribution refers to the amount of contribution of various resources used to participate in the FL process. Resources include data resources, computing resources, communication resources, etc. User reputation refers to the honesty of participating in the training process.
\par

\textbf{2) Reduce cost (C).} Minimize the cost, or within the budget. It is mainly to plan the problem into a cost model and find the optimal solution on this basis.
\par

\textbf{3) Reduce efficiency and energy consumption (E).} Reduce the resource consumption of servers and participants, including computing resources, communication resources, computing time, etc. Improve the efficiency of calculation.
\par

\textbf{4) Improve system security (S).} Set up corresponding system protection mechanisms to avoid malicious attacks by users and the influence of dishonest users on the system.
\par

\textbf{3. System elements:} Design the incentive mechanism by considering which elements of the system.
\par

\textbf{4. Incentive function:} Design the corresponding incentive function through system elements.
\par

\textbf{5. Interaction mechanism:} Which interaction mechanism is used to realize the dynamic interaction between the incentive subject and the incentive object.
\par

\textbf{6. Reward Method (RM):} The way of rewarding clients. There are two reward methods:
\begin{itemize}[leftmargin=*]
	\item \textbf{Money (M).} Reward participants through money.
	\item \textbf{Non-money (NM).} Reward participants through non-monetary means, such as points.
\end{itemize}

\par

We conducted statistical analysis on the scenes, optimization goals, system elements, incentive functions, interaction mechanisms, reward methods, and incentive effects of the incentive mechanism for FL, as shown in Table \ref{table_3-10}.

\begin{table*}[htbp]\scriptsize
	\caption{Statistics on the incentive mechanism for FL}
	\label{table_3-10}
	\centering
	\begin{tabular}{c|m{2.5cm}<{\centering}|m{0.5cm}<{\centering}|m{4.9cm}<{\raggedright}|m{1.8cm}<{\centering}|m{1.4cm}<{\centering}|c|m{3cm}<{\centering}}
		\hline
		\textbf{Article} & \textbf{Scene} & \textbf{OG} & \textbf{System element} & \textbf{Incentive function} & \textbf{Interaction mechanism} & \textbf{RM} & \textbf{Incentive effect} \\ \hline
		\cite{DBLP:conf/aies/0001LLCCWNY20} & Clients make alliance joining decisions based on returns, and the alliance quantifies each client's revenue to achieve long-term system benefits & Q & \tabincell{l}{1. $t$ current training rounds \\2. $q(t)$ the resource quality of the client, \\including computing resources, data\\ resources, etc. \\3. $B(t)$ the share of the client's current \\return \\4.  $c(t)$ client participation cost \\5. $Y(t)$ the difference between the return \\received by the client so far and the\\ expected return \\6. $Q(t)$ time queue for the client to wait \\for all returns} & Maximize profit 
		$\begin{array}{l}
			\max G = \\
			f(t,q(t),\\
			B(t),c(t),\\
			Y(t),Q(t))
		\end{array}$
		& payoff-sharing scheme & M & Compared with five methods (Shapley/ Individual Union/ Equal/ Linear), the data quality weighting result is 7\% higher than others, and the income is 1.5\% higher than others. \\ \hline
		\cite{lim2020hierarchical} & Collaborative machine learning among multiple model owners in mobile networks & Q & \tabincell{l}{1. $d$ quantity of client data \\2. $q$ quality of client data \\3. $s$ federated scale \\4. $c$ cost of server coordination training} & Maximize profit $\max G = f(d,q,s,c)$& Contract theoretic hierarchical incentive mechanism & M & Based on contract theory, forming a consortium that can achieve goal of maximizing profits, the scheme has a total profit of more than 10,000. \\ \hline
		\cite{DBLP:conf/globecom/PandeyTBTHH19} & Crowdsourcing scenes for machine learning & Q,E & \tabincell{l}{1. $\theta $ accuracy of the client model \\2. $com(\theta )$ communication cost of client \\3. $cmp(\theta )$ computing cost of the client \\4.
			$\varepsilon $ accuracy of the global model \\5. $x(\varepsilon )$ number of server-side iterations} & Maximize utility 
		$\begin{array}{l}
			\max G = f(\theta ,\\
			com(\theta ), \varepsilon ,
			\\cmp(\theta ) ,x(\varepsilon ))
		\end{array}$
		& Two-stage Stackelberg game & M & The scheme has higher performance than the heuristic method and can obtain up to 22\% reward. \\ \hline
		\cite{DBLP:conf/bigcom/BaoSXHH19} & Commercialized collaborative training model market & Q,S & \tabincell{l}{1. $ep$ expected price of server \\2. $ei$ expected income of server \\3. $c$ estimated cost of server \\4. $q$ quality of client model} & Maximize profit 
		$\begin{array}{l}
			\max G = \\
			f(ep,ei,
			\\c,q)
		\end{array}$
		& Blockchain-based incentive mechanism & M & This scheme can encourage clients to honestly participate in training and detect client misconduct. \\ \hline
		\cite{DBLP:conf/apwcs2/KangXN0LK19} & Task release and claim scenes in mobile networks & Q,E & \tabincell{l}{1. $\theta $ accuracy of the client model \\2. $T(\theta )$ total time of one iteration \\3. $m$ quality category of client data \\4. $cmp$ computing cost of the client \\5. ${T_{\max }}$ specified training time of server} & Maximize profit 
		$\begin{array}{l}
			\max G = \\
			f(\theta ,T(\theta ),\\
			m,cmp,{T_{\max }})
		\end{array}$
		& Contract theoretic incentive mechanism & M & The mission publisher profit of this scheme is higher than the Stackelberg game model. \\ \hline
		\cite{DBLP:journals/iotj/KangXNXZ19} & Task release and claim scenes in mobile networks & Q, E, S & \tabincell{l}{1. $\theta $ accuracy of the client model \\2. $T(\theta )$ total time of one iteration \\3. $m$ quality category of client data \\4. $cmp$ computing cost of the client \\5. ${T_{\max }}$ specified training time of server}  & Maximize profit 
		$\begin{array}{l}
			\max G = f(\theta \\
			,T(\theta ),  m,\\
			cmp,{T_{\max }})
		\end{array}$
		& Incentive mechanism combining reputation and contract theory & M & This scheme accurately evaluates reputation, and incentivizes high-reputation clients to participate in FL. Finally, the ACC is increased by 0.1. \\ \hline
		\cite{DBLP:conf/bigdataconf/0013SLY19} & Power demand response, meet the balance of power supply and demand & C & \tabincell{l}{1. $b$ bid price of client \\2. $q$ interrupt response value of client} & Minimize costs $\min G = f(p,q)$& Multi-attribute sealed auction game & NM & This scheme increases the probability of SMEs and residents selected, and reduces the total cost of power company. \\ \hline
		\cite{DBLP:conf/bigdataconf/SongTW19} & Profit distribution scene in collaborative machine learning & Q & \tabincell{l}{1. performance of client model} & None & payoff-sharing scheme & M & Compared with other measures of contribution, the speed of this scheme is increased by 2-100 times. \\ \hline
		\cite{zhan2020learning} & Edge node collaborative training scene & Q, C, E &\tabincell{l}{1. $d$ quantity of client data \\2. $cmp$ computing cost of the client \\3.$com$ communication cost of client}  & Maximize the profit of edge nodes 
		$\begin{array}{l}
			\max G = f(d,
			\\cmp,com)
		\end{array}$
		& Deep reinforcement learning-based incentive mechanism & M & Both the server pricing strategy and the node pricing strategy of this scheme can converge to the Stackelberg game level. \\ \hline
		\cite{liu2020fedcoin} & Blockchain-based FL peer-to-peer payment system & Q &\tabincell{l}{1. quality of client data}  & None & Shapley value based on profit distribution & M & The scheme builds an incentive system for FL. \\ \hline
		\cite{zeng2020fmore} & Edge node collaborative training scene & Q, C, E & \tabincell{l}{1. $q$ quality of client resources, measured \\by computing power, data quality, and CPU \\cycles \\2. $p$ pay of client} & Maximize profit $\max G = f(q,p)$& K-winner multi-dimensional procurement auction mechanism & M & This scheme accelerates the speed of training by reducing 51.3\% training rounds, and improves ACC by 28\%. \\ \hline
		\cite{jiao2020toward} & Wireless FL services market & Q, E & \tabincell{l}{1. $c$ the total cost of data collection, \\calculation and communication of client \\2. $\sigma $ distribution of client data \\3. $b$ bid price of client \\4. $d$ quantity of client data} & Maximize profit 
		$\begin{array}{l}
			\max G = f(c,
			\\\sigma ,b,d)
		\end{array}$
		& Automated deep reinforcement learning based auction mechanism & M & This scheme is higher than the other two methods (reverse multi-dimensional auction mechanism, baseline). \\ \hline
		\cite{khan2019federated} & Edge node collaborative training scene & Q, E & \tabincell{l}{1. $cmp$ computing cost of the client \\2. $com$ communication cost of client \\3. $\theta $ accuracy of the client model} & Maximize training performance 
		$\begin{array}{l}
			\max G = f(\\
			cmp, com,\theta )
		\end{array}$
		& Stackelberg game & M & None \\ \hline
	\end{tabular}
\end{table*}

From Table \ref{table_3-10}, the following conclusions can be drawn:
\begin{itemize}[leftmargin=*]
	\item From the perspective of optimization goals and effects, 12 articles (92\%) focused on the quality of the trained model. They improved the training quality of the global model through the incentive mechanism. In addition, seven articles (54\%) also studied the issue of efficiency and energy consumption, reducing the overall energy consumption of servers and clients through incentive mechanisms. For example, article \cite{zeng2020fmore} speeds up training through incentive mechanisms. Eight papers (62\%) finally achieved the maximization of FL benefits.
	\item From the perspective of the system elements considered, the client’s resource quality (including model quality, data quality, and data quantity) is currently considered the most, followed by communication cost and calculation cost. The system elements considered are related to the optimization goal. The optimization goal is to improve the quality of the model, resulting in that the system element considered is the resource quality of the client; The optimization goal is to reduce cost control and energy consumption, resulting in that the system elements considered are communication and computing costs.
	\item From the perspective of reward methods, most articles motivate participants by money, and the only article \cite{DBLP:conf/bigdataconf/0013SLY19} motivate participants by non-monetary means, through reward points to increase the probability of being selected, and further encourage more small and medium-sized users to participate. Participants aim to obtain monetary rewards, which is easy to cause fraud. And a single reward method cannot fully invoke the enthusiasm of the participants. At present, there is no incentive mechanism that combines multiple reward methods, such as combining monetary incentives with non-monetary incentives, which can become one of the future research directions.
\end{itemize}

\subsection{RQ5: Can FL achieve similar learning effects to non-FL on the same dataset? What are the conditions when FL and non-FL results are similar?}
\begin{figure}[htbp]
	\centering
	\includegraphics[width=3in]{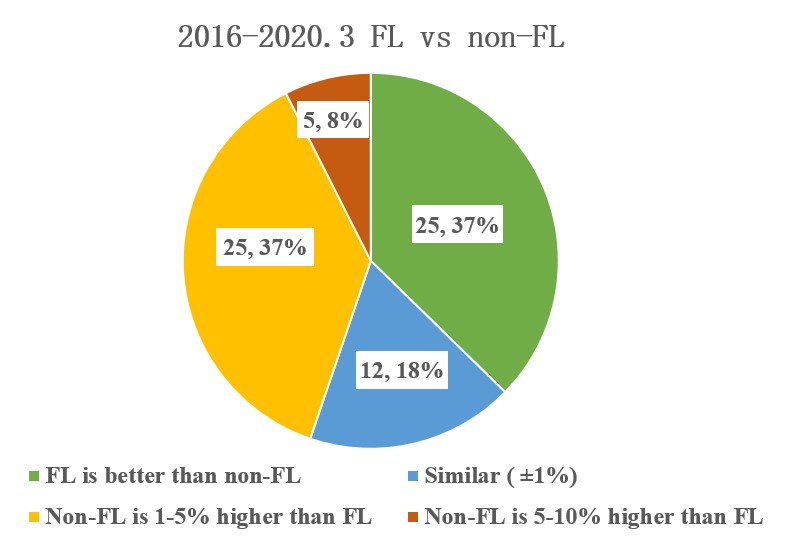}
	\caption{  (2016-2020.3) Statistics of comparison of effects of FL and non-FL}
	\label{fig_3-6}
\end{figure}

We focused on whether FL can replace non-FL. In response to this problem, we conducted statistics on the effects of FL and non-FL, as shown in Fig. \ref{fig_3-6}. A total of 67 articles were counted. 62 (92\%) of the articles showed that the FL effect is not much different from the non-FL effect or the FL effect is better than the non-FL effect. There are five papers (8\%) where there is too much difference between the non-FL and the FL effect, indicating that FL can achieve similar learning effects to non-FL.

In addition to factors such as aggregation algorithms, incentive mechanisms, and improved algorithms, we analyze other factors that affect the effect of FL. For the same data set and the same client algorithm, we counted the best effect, the worst effect, the data set used, and the factor of effect. The statistical result is shown in Table \ref{table_3-11}. The performance indicators in the table are all ACC. In the table, IID (Independent Identically Distributed) means that the values at any time in the random process are random variables. If these random variables follow the same distribution and are independent of each other, these random variables are independent and identically distributed.

\begin{table*}[htbp]
	\small
	\caption{Comparison of effect of FL under the same data set and the same client algorithm}
	\label{table_3-11}
	\centering
	\begin{tabular}{m{2cm}<{\centering}|m{2cm}<{\centering}|m{2cm}<{\centering}|m{2cm}<{\centering}|m{2cm}<{\centering}|m{5cm}<{\centering}|m{1cm}<{\centering}}
		\hline
		\textbf{FL algorithm} & \textbf{Best result} & \textbf{Worst result} & \textbf{Diffe- rence} & \textbf{Data set} & \textbf{Factors affecting results} & \textbf{Article} \\ \hline
		\multirow{4}{*}{CNN} & 0.7473 & 0.6426 & 0.1047 & USC-HAD \cite{USC-HADData} & The training results without noise data are better than those with noisy data & \cite{chen2020focus} \\ \cline{2-7} 
		& 0.99 & 0.952 & 0.038 & minist \cite{ministData} & IID data training results are better than non-IID data training results & \cite{alotaibi2018wisdom,DBLP:conf/icip/0003HWZS19} \\ \hline
		MLP & 0.9816 & 0.7 & 0.2816 & minist \cite{ministData} & The training results of homogeneous data are better than those of heterogeneous data & \cite{DBLP:conf/middleware/NilssonSUGJ18,DBLP:conf/icml/YurochkinAGGHK19} \\ \hline
	\end{tabular}
\end{table*}
\begin{table*}[htbp]
	\small
	\caption{ Statistics of the comparison between the effects of FL and non-FL with the five client algorithms}
	\label{table_3-12}
	\centering
	\begin{tabular}{m{2.3cm}<{\centering}|m{1.4cm}<{\centering}|m{4cm}<{\centering}|m{3cm}<{\centering}|m{3cm}<{\centering}|m{1.8cm}<{\centering}}
		\hline
		\multirow{2}{*}{\shortstack{\textbf{Comparison of } \\\textbf{FL and non-FL }\\\textbf{effects}}} & \multirow{3}{*}{\shortstack{\textbf{Client}\\ \textbf{algorithm}}} & \multirow{2}{*}{\textbf{Factor for improvement}} & \multicolumn{2}{c|}{\shortstack{\textbf{Effect (The default indicator} \\\textbf{is ACC, special will be marked)}}} & \multirow{2}{*}{\textbf{Article}} \\ \cline{4-5}
		&  &  & \textbf{FL} & \textbf{Non-FL} &  \\ \hline
		\multirow{9}{*}{\shortstack{FL is better \\than non-FL}} & \multirow{3}{*}{CNN} & Improve aggregation algorithm & 0.8753 & 0.7529 &\cite{DBLP:conf/iclr/WangYSPK20} \\ \cline{3-6} 
		&  & Choose high-quality clients & \multicolumn{2}{c|}{FL effect is 25\% higher than non-FL} & \cite{tuor2020data} \\ \cline{3-6} 
		&  & Add transfer learning & \multicolumn{2}{c|}{FL error rate is 17.39\% lower than non-FL} & \cite{liu2019federated} \\ \cline{2-6} 
		& \multirow{2}{*}{RNN} & Improve aggregation algorithm & 0.4907 & 0.4606 & \cite{DBLP:conf/iclr/WangYSPK20} \\ \cline{3-6} 
		&  & (Not improved) & \multicolumn{2}{c|}{FL interruption rate is 16\% lower than non-FL} & \cite{DBLP:conf/globecom/ChenSSLY19} \\ \cline{2-6} 
		& DNN & (Not improved) & 0.9651 & 0.9351 & \cite{DBLP:conf/soict/ZhaoCWTY19}\\ \cline{2-6} 
		& \multirow{2}{*}{MLP} & Anonymous hybridization of transmitted data & 0.83 & 0.809 & \cite{cui2019federated} \\ \cline{3-6} 
		&  & Add multi-task learning & 0.56 & 0.44 & \cite{corinzia2019variational} \\ \cline{2-6} 
		& LR & Improve aggregation algorithm & 0.801 & 0.766 & \cite{DBLP:journals/corr/abs-1910-06378} \\ \hline
		\multirow{8}{*}{Similar ($\pm$ 1\%)} & CNN & \multirow{3}{*}{(Not improved)} & \multicolumn{2}{c|}{\multirow{14}{*}{(Not promoted)}} & \cite{DBLP:conf/miccai/ShellerREMB18,tobis2019federated} \\ \cline{2-2} \cline{6-6} 
		& RNN &  & \multicolumn{2}{c|}{} & \cite{DBLP:conf/globecom/HuGLM18,DBLP:conf/hotedge/LuYS19,DBLP:conf/icdcs/NguyenMMFAS19,ramaswamy2019federated,hard2018federated,DBLP:journals/imwut/FengRSGL20} \\ \cline{2-2} \cline{6-6} 
		& DNN &  & \multicolumn{2}{c|}{} & \cite{DBLP:conf/soict/ZhaoCWTY19,DBLP:journals/access/YeYPH20} \\ \cline{2-3} \cline{6-6} 
		& \multirow{3}{*}{MLP} & Improve aggregation algorithm & \multicolumn{2}{c|}{} & \cite{DBLP:conf/icml/YurochkinAGGHK19} \\ \cline{3-3} \cline{6-6} 
		&  & Improved algorithm & \multicolumn{2}{c|}{} & \cite{liu2018fadl,corinzia2019variational} \\ \cline{3-3} \cline{6-6} 
		&  & (Not improved) & \multicolumn{2}{c|}{} & \cite{choudhury2019differential} \\ \cline{2-3} \cline{6-6} 
		& \multirow{2}{*}{LR} & Improve aggregation algorithm & \multicolumn{2}{c|}{} & \cite{yang2019quasi} \\ \cline{3-3} \cline{6-6} 
		&  & (Not improved) & \multicolumn{2}{c|}{} & \cite{pfohl2019federated} \\ \cline{1-3} \cline{6-6} 
		\multirow{5}{*}{\shortstack{Non-FL is 1-5\% \\higher than FL}} & CNN & \multirow{5}{*}{(Not improved)} & \multicolumn{2}{c|}{} & \cite{chen2020federated,yoon2020federated,qi2020fedrec} \\ \cline{2-2} \cline{6-6} 
		& RNN &  & \multicolumn{2}{c|}{} & \cite{DBLP:journals/corr/abs-1910-12191,DBLP:journals/imwut/FengRSGL20} \\ \cline{2-2} \cline{6-6} 
		& DNN &  & \multicolumn{2}{c|}{} & \cite{DBLP:conf/ispa/SozinovVG18} \\ \cline{2-2} \cline{6-6} 
		& MLP &  & \multicolumn{2}{c|}{} & \cite{choudhury2019differential,ciftler2020federated,li2020multi} \\ \cline{2-2} \cline{6-6} 
		& LR &  & \multicolumn{2}{c|}{} & \cite{DBLP:conf/ispa/SozinovVG18,choudhury2019differential,pfohl2019federated} \\ \cline{1-3} \cline{6-6} 
		Non-FL is 5-10\% higher than FL & DNN & (Not improved) & \multicolumn{2}{c|}{} & \cite{DBLP:journals/access/Wang00G19,samal2019time} \\ \hline
	\end{tabular}
\end{table*}

It shows that the effect of FL on the same data set and the same client algorithm will be different due to the division, quality, and heterogeneity of clients’ data.
\par
For the client-side algorithms, we focused on the statistical analysis of the effects of FL and non-FL on CNN, RNN, DNN, MLP, LR. We only select the articles containing the comparison of FL and non-FL effects. For the five client algorithms CNN, RNN, DNN, MLP, and LR, we compared the effects of FL and non-FL, the factors for the improvement, and the effects. In the effect column, the default indicator is ACC. If there are special circumstances, it will be marked separately. The statistical result is shown in Table \ref{table_3-12}.

It can be seen from Table \ref{table_3-12} that different articles have different FL and non-FL effect (±1\%) differences between 1-5\% due to different data sets. But due to the new aggregation algorithms or improved algorithms, the effect difference can be similar (±1\%), and the FL effect can even be better than the non-FL effect.
\par
In short, in the current 62 (92\%) articles, the difference between the model effects of FL and non-FL is within 5\%, of which 25 (37\%) articles, the difference between the model effects of FL and non-FL is 1-5\%. FL achieve similar learning effects to non-FL in various applications.

According to the study of other articles\cite{kairouz2019advances, DBLP:conf/iccd/DuanLCTRQL19}, the effect of FL is mainly related to the distribution and the amount of data. Two important facts are summarized as follows.
\begin{enumerate}
\item
When the data is independent and identically distributed, the learning results of FL and non-FL are similar. For instance, the work \cite{DBLP:conf/hotedge/LuYS19,chen2020federated,DBLP:conf/ispa/SozinovVG18,choudhury2019differential} showed that when the training data is independent and identically distributed (IID), the difference between FL and non-FL is within 3\%. If the amount of data on each client is small, FL's effect is better than that of non-FL because FL expands the number of IID data samples \cite{DBLP:conf/mobicom/IckinVF19, bakopoulou2019federated, roy2019braintorrent, liang2019federated, chen2020fl, suzumura2019towards}.
\item
When the training data is unevenly distributed, FL may not achieve the effect of non-FL \cite{DBLP:conf/miccai/ShellerREMB18, DBLP:conf/globecom/HuGLM18, DBLP:journals/imwut/FengRSGL20, liu2018fadl, corinzia2019variational, chen2020federated, DBLP:journals/corr/abs-1910-12191, DBLP:conf/ispa/SozinovVG18}. The review article \cite{ kairouz2019advances} claimed that in the case of uneven distribution of training data, if the amount of data on a client is small, the non-FL model may be better than the FL model trained using data on multi-clients.
\end{enumerate}

In addition, some works have shown that the effect of FL is also related to the encryption algorithms \cite{pfohl2019federated,qi2020fedrec,DBLP:journals/imwut/FengRSGL20,choudhury2019differential,DBLP:journals/tii/LuHDMZ20,DBLP:conf/miccai/LiMXRHZBCOCF19,DBLP:journals/cem/LiSM20}. If the encryption strength increases, the information loss on the data will be worse. So that the effect of FL decreases. The trade-off between data privacy protection and the accuracy of the trained model is still inevitable.

It is still an open question how the effect of FL is related to other factors. There is no more work exploring it.

\section{Federated Learning Application Framework}
\subsection{Overall application framework of FL}

\begin{figure}[htbp]
	\centering
	\includegraphics[width=3.5in]{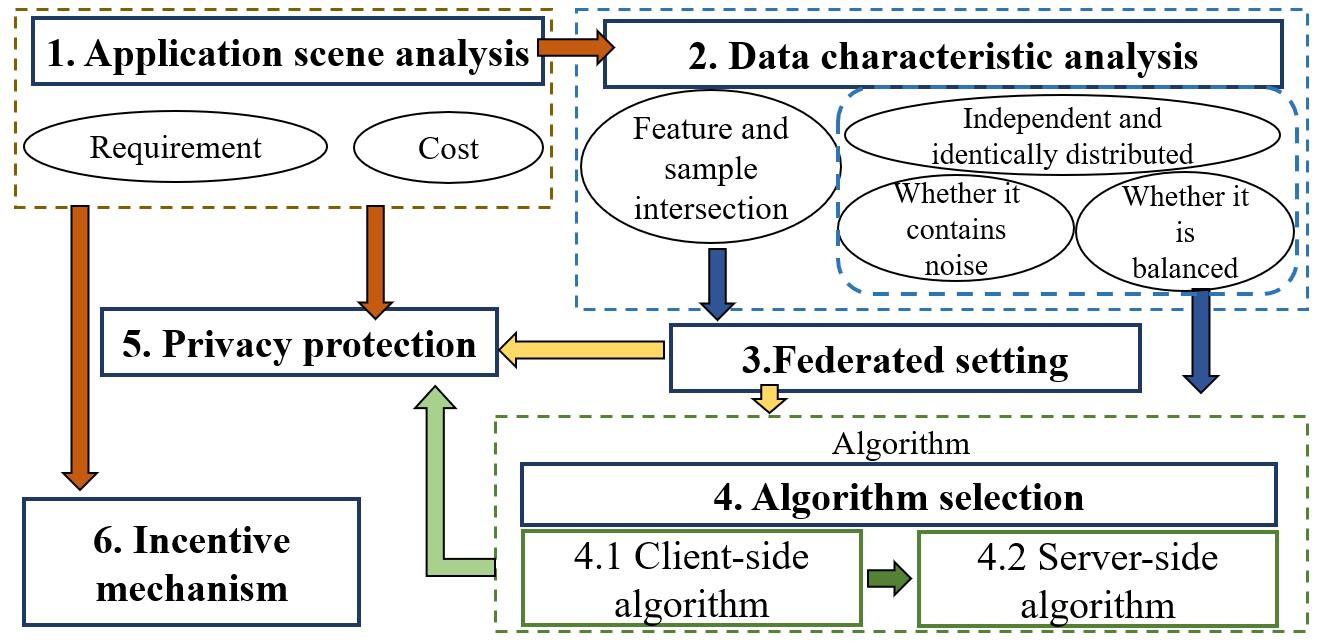}
	\caption{ Application framework of FL}
	\label{fig_4-1}
\end{figure}

In this section, through the analysis of the statistical results of the previous chapters, we give the overall application framework of FL. The application framework of FL is shown in Fig. \ref{fig_4-1}.

\begin{enumerate}[leftmargin=*]
	\item Application scene analysis. First, start from the application scene of FL, analyze the potential demands contained in the scene, such as requirement and cost.
	\item Data characteristics analysis. Analyze the characteristics of the intersection of features and samples and the characteristics of the data.
	\item Federated setting. The intersection of the characteristics of each client and the sample will determine the federated setting. For the specific method of selecting the federated setting, please refer to the detailed introduction of the federated setting in section 2.1.
	\item Algorithm selection. The federated setting will further affect the choice of federated algorithm. Other characteristics of the data, such as whether it is independent and identically distributed, whether it contains noise, whether it is balanced, etc., will also affect the choice of the FL algorithm. Section 3.2 presents the current client-side and server-side algorithms in the four federated settings: horizontal, vertical, transfer, and reinforcement. Section 3.2 also provides client-side and server-side algorithms for different characteristics of the data. In Section 4.2, we further give a framework for selecting client-side and server-side algorithms for FL for optimization purposes.
	\item Privacy protection. When designing a scheme to improve privacy, you need to consider the scenes, federated setting, and selection of federated algorithms. Section 3.5 introduces the current privacy protection measures in detail. In Section 4.3, we give a framework for selecting privacy protection mechanisms.
	\item Incentive mechanism. The results of the scene analysis will determine how the incentive mechanism is designed. Section 3.3 gives the current articles on the incentive mechanism for FL and introduces how to design the corresponding incentive mechanism in each document under practical scenes. Section 4.4 gives specific suggestions for the design process of the incentive mechanism.
\end{enumerate}

\subsection{FL algorithm selection}
\begin{figure}[htbp]
	\centering
	\includegraphics[width=3.5in]{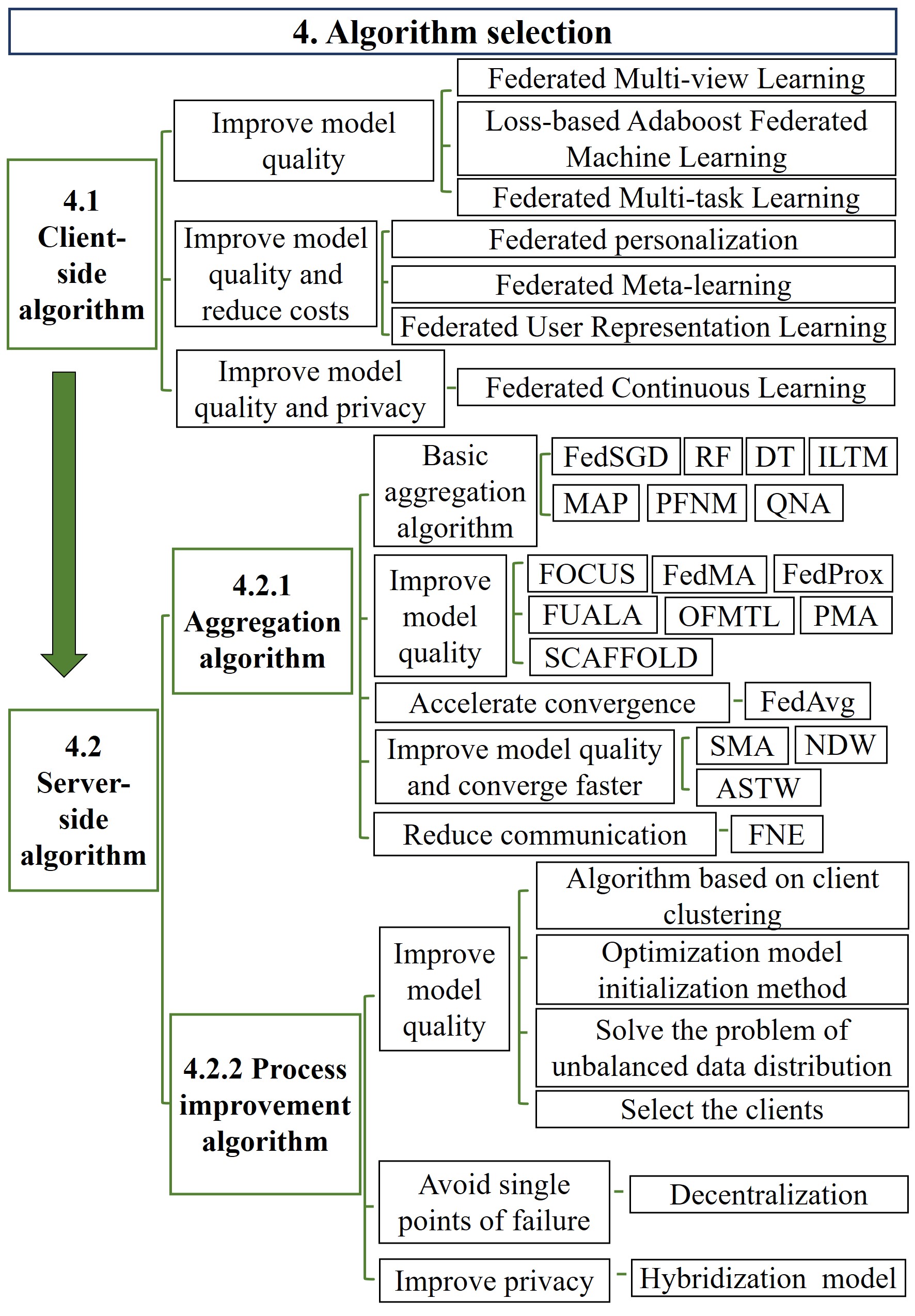}
	\caption{FL algorithm selection framework}
	\label{fig_4-2}
\end{figure}

In Section 3.2, we have performed a statistical analysis of the client-side and server-side algorithms of FL. In Fig.\ref{fig_4-2}, we give a framework for selecting FL client-side and server-side algorithms.
\par
\par
For the client-side algorithm, we can design from three aspects: improving model quality, reducing communication costs, and improving privacy. The available methods or techniques have been introduced in detail in Chapter 3.
\par

For the server-side algorithm, we can design from two aspects: the aggregation algorithm and process improvement algorithm. For the aggregation algorithm, we can design from four objectives: basic algorithms, improving model quality, accelerating convergence, and reducing communication; for process improvement algorithms, we can design from three objectives: improving model quality, avoiding single points of failure, and improving privacy. The available methods or techniques have been introduced in detail in section 3.2.

\subsection{FL privacy protection scheme selection}
\begin{figure}[htbp]
	\centering
	\includegraphics[width=3in]{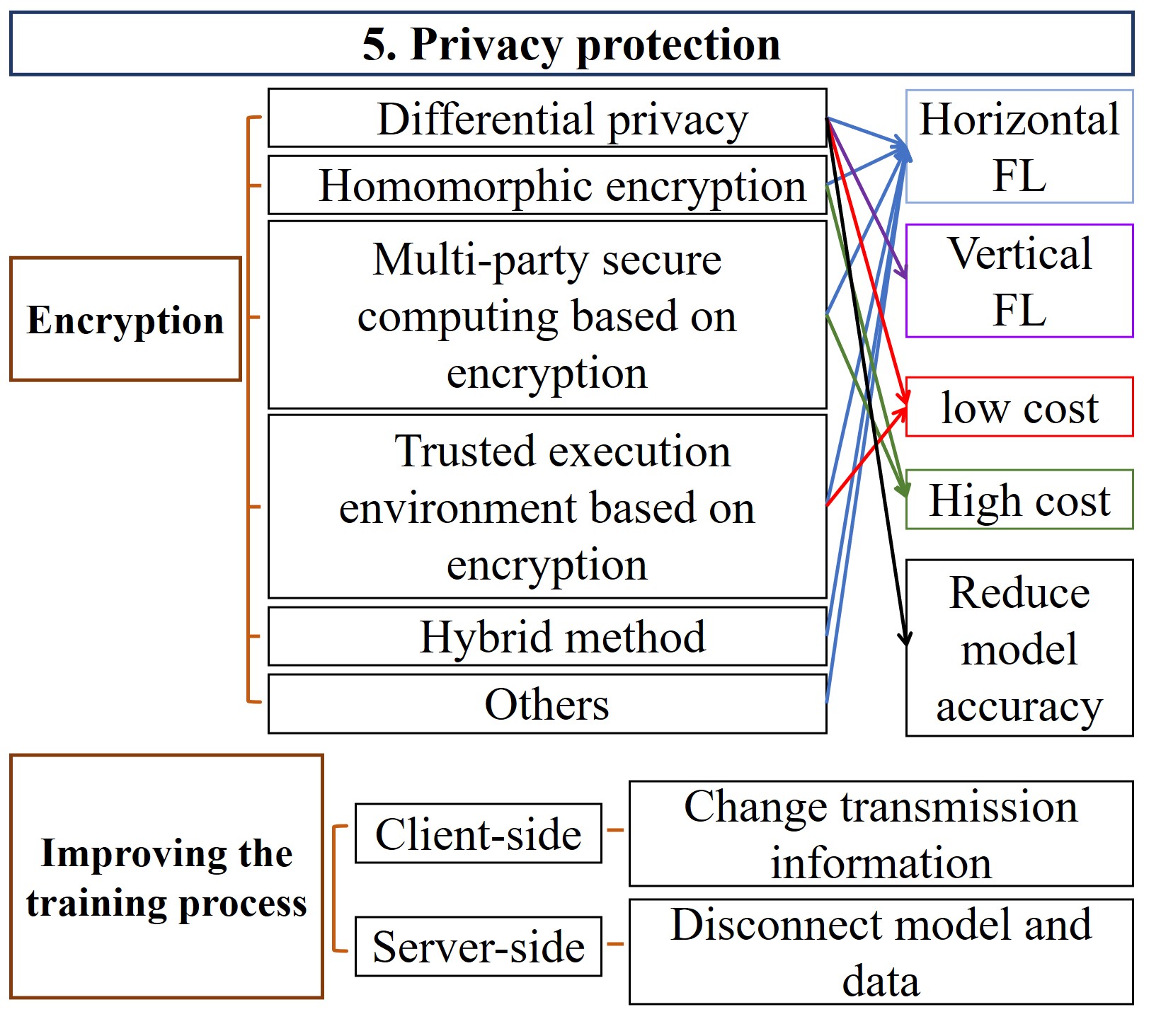}
	\caption{FL privacy protection program framework}
	\label{fig_4-3}
\end{figure}

In Section 3.3, we have introduced various privacy protection technologies in detail. After our refining and analysis, we give the selection framework of the FL privacy scheme in Fig. \ref{fig_4-3}. When choosing a privacy scheme for FL, we should choose a reasonable privacy protection scheme for FL based on practical scenes.

\subsection{FL incentive mechanism design process}
\begin{figure}[htbp]
	\centering
	\includegraphics[width=3.3in]{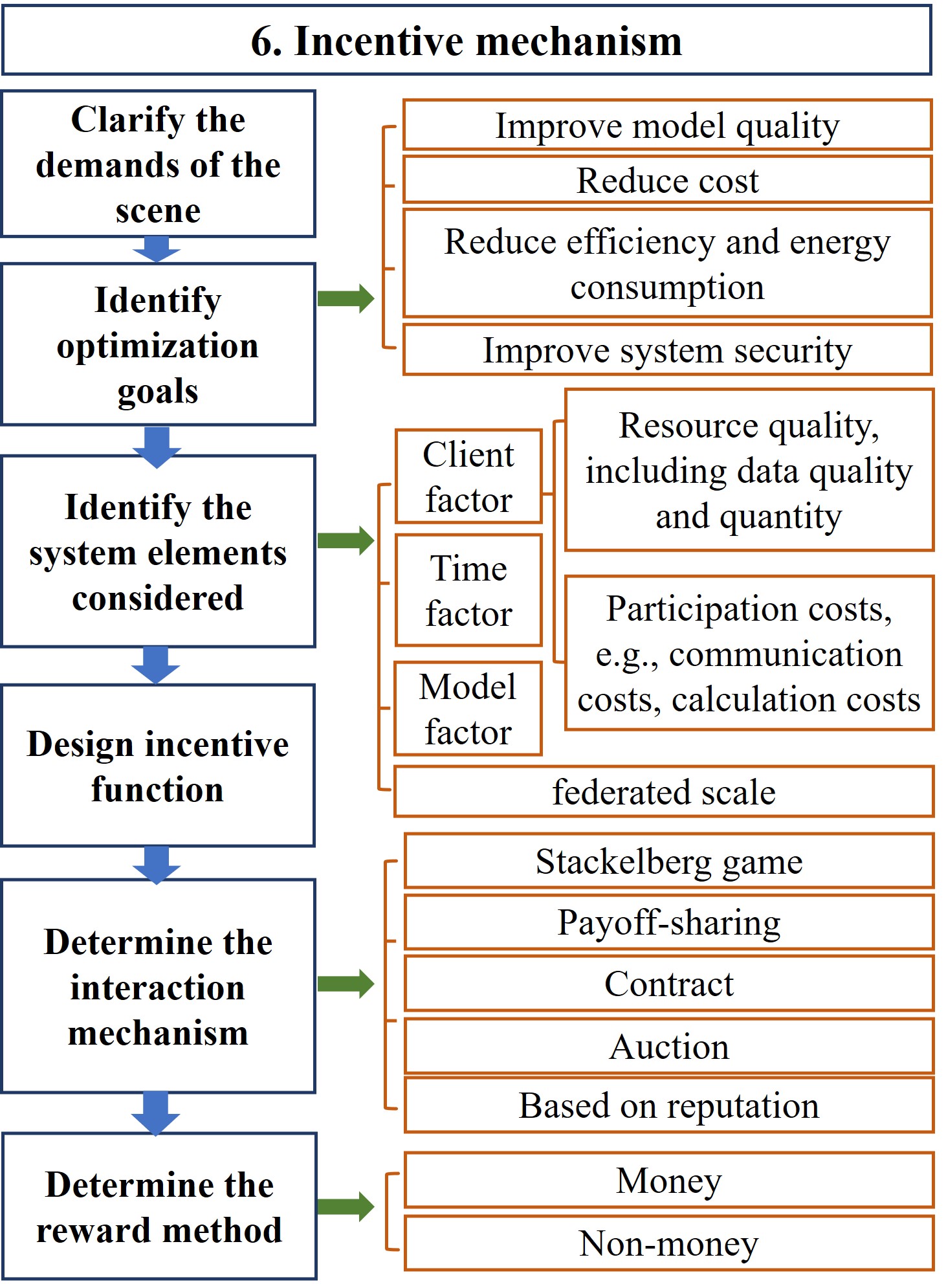}
	\caption{ FL incentive mechanism design process}
	\label{fig_4-4}
\end{figure}

In Section 3.4, we have conducted a detailed statistical analysis of the current incentive mechanism for FL. We propose the design process of the incentive mechanism, as shown in Fig. \ref{fig_4-4}.

\section{Threats to Validity}
The main threat to the validity of this systematic literature review is related to the completeness of literature selection and data extraction accuracy.

\subsection{Completeness of Literature Selection}
To ensure the completeness of literature selection, in section 2.2, we determined the search keywords after many attempts, adjusted the search terms according to the different search functions and search refinement process provided by various online databases. We adopted a systematic and fair literature selection strategy by establishing a three-stage document selection process. Each stage has different selection criteria, and each step records the reasons for literature exclusion. For literature with uncertainties, we argued and made decisions. We made sure to include rather than exclude literature to a large extent.

\subsection{Accuracy of Data Extraction}
Subjectivity may lead to inaccurate data extraction. The four authors of this systematic literature review participated in the data extraction process. The first three authors analyzed all selected documents, and the other author supervised the whole process. Besides, in our systematic literature review, we also deal with this problem by developing a data extraction form. Initially, we conducted a preliminary data extraction to fill a table sample regarding the RQs. Each time, we randomly selected ten documents and refined the data extraction table based on these ten documents' data extraction results. This process was repeated until a complete data extraction form was formed.

\section{Conclusions}
This systematic literature review focuses on 147 articles related to 5 RQs about FL: (1) the research and application trends of FL; (2) the quality of model design and training; (3) the quality of data privacy protection mechanism; (4) the quality driven by the incentive mechanisms; (5) the effect comparison between FL and non-FL. After conducting the review, we get some important conclusions, listed hereafter.
\par
\textbf{(1) The research and application trends of FL}
\par
The current research trend of FL is still in its infancy. As a new research direction, there are still many gaps in both theory and application. Therefore, many research results have emerged in the short term. Among the research scenes of FL, edge computing \& IoT (33\%), healthcare (29\%), Urban Computing \& Smart City (17\%) have more applications than other fields. There are a few works on applying FL to industrial manufacturing and other fields. However, the data island problem is obvious in these fields, and there is a strong need for FL. How to apply FL to the problems in these fields and how to solve them are open questions.
\par
\textbf{(2) The quality of model design and training}
\par
In current research works, the client-side and server-side of FL usually use the same type of machine learning model. In very few articles  (2/88), the client-side model is different from the server-side model. The models on both sides have a similar parameter structure, or it is possible to convert the client-side model to the server-side one. More works focused on the design and training of FL models used in horizontal FL. There are many open questions in vertical FL, federated transfer learning, and federated reinforcement learning. 

According to our review, the client-side algorithm like Federated User Representation Learning \cite{bui2019federated}, Multi-view FL \cite{huang2019Iterative}, and Multi-task FL \cite{corinzia2019variational} are significantly effective for improving the learning quality. The server-side aggregation algorithms are divided into model aggregation and parameter aggregation. The server-side algorithm like FedProx\cite{DBLP:journals/corr/abs-1812-06127}, SMA\cite{DBLP:journals/access/YeYPH20} and FOCUS\cite{chen2020focus} are significantly effective for improving the learning quality. Concerning the FL's training process, the algorithm based on client clustering \cite{sattler2019clustered} is more effective than other algorithms for improving the learning quality.
\par
\textbf{(3) The quality of data privacy protection mechanism}
\par
We can improve the privacy of FL from two aspects. One is to use encryption algorithms in data transmission and training. The other is to weaken the connection between the model and the original data in the training process.
\par
Currently, most of the encryption algorithms used for FL are borrowed from distributed computing. Most algorithms have been theoretically proven to achieve data privacy protection. Most algorithms are applied to horizontal FL. A future research direction is to design suitable encryption algorithms for the vertical FL, federated transfer learning, and federated reinforcement learning.
\par
Improving the quality of privacy protection from the client-side or server-side in FL is still an open question. There are a limited number of works in this direction. Many relevant problems need to be studied.
\par
\textbf{(4) Quality driven by incentive mechanism}
\par
Most articles (12/13) focused on the quality of the model. They improved the training quality of the global model by using some incentive mechanism. Most works have considered the client's resource quality (including model quality, data quality, and data quantity), communication cost, and calculation cost. About the rewards in the incentive mechanism, most works use monetary incentives. A single incentive measure cannot well invoke the enthusiasm of participants. Nevertheless, no incentive mechanism combines multiple reward methods, such as combining monetary incentives with non-monetary incentives, which can become a future research direction.
\par
\textbf{(5) Effect comparison between FL and non-FL}
\par
Among the 62 articles (92\%), the difference of learning quality between FL and non-FL is within 5\%, among which 25 (37\%) articles differ by 1- 5\%. Although FL does not achieve the same learning quality as non-FL in some cases, it achieved well privacy protection. According to the application results in the literature, a difference value within 1-5\% are generally acceptable. 

The known factors differing the learning quality between FL and non-FL include the IID of the training data, the amount of data, and the encryption algorithm. It is an open question of how other factors affect and differ the learning quality between FL and non-FL.

\par
In conclusion, we conducted a systematic literature review related to FL. We place the focus on the primary factors affecting the FL model quality, covering the model design and training, the privacy protection mechanism, and the incentive mechanism. We are also interested in the research and application trends of FL and the effect comparison between FL and non-FL. The practitioners frequently ask the latter because many people worry that achieving privacy protection needs compromising the learning quality. Based on the answers to the RQs, we propose an FL application framework. We provide some suggestions for configuring the options in the critical steps of design to improve FL model quality.


%



\section*{Acknowledgment}
This work was supported by the National Key Research and Development Program of China Grant No. 2019YFB1703903, National Natural Science Foundation of China Grant No. 61732019 and No. 61902011.




\bibliography{Ref}
%

\bibliographystyle{unsrt}

\end{document}